\documentclass[11pt]{article}

\usepackage[preprint]{acl-onecolumn}

\usepackage{iftex}
\ifPDFTeX
  \usepackage{times}
  \usepackage[T1]{fontenc}
  \usepackage[utf8]{inputenc}
  \DeclareRobustCommand{\ph}[1]{#1}
  \DeclareRobustCommand{\phA}{A}
\else
  \usepackage{fontspec}
  \newfontfamily\phoenicianfont{NotoSansPhoenician-Regular.ttf}[Path=./]
  \DeclareRobustCommand{\ph}[1]{\ifmmode\text{\phoenicianfont #1}\else{\phoenicianfont #1}\fi}
  \DeclareRobustCommand{\phA}{\ph{\char"10900}}
\fi
\usepackage{microtype}

\usepackage{amsmath, amssymb, amsfonts}
\usepackage{dsfont}
\usepackage{bookmark}
\usepackage{graphicx}
\usepackage{booktabs}
\usepackage{siunitx}
\usepackage{enumitem}
\usepackage{float}
\newsavebox\maxwidthboxbox
\newcommand{\maxwidthbox}[1]{\sbox\maxwidthboxbox{#1}\ifdim\wd\maxwidthboxbox>\linewidth\resizebox{\linewidth}{!}{\usebox\maxwidthboxbox}\else\usebox\maxwidthboxbox\fi}

\usepackage{tikz}
\usepackage{pgfplots}
\pgfplotsset{compat=1.18}
\usetikzlibrary{arrows.meta, positioning, shapes.geometric, shapes.misc, calc, fit, backgrounds}

\usepackage{algorithm}
\usepackage{algpseudocode}

\usepackage[nameinlink,noabbrev]{cleveref}

\newcommand{\canvas}[1]{\texttt{#1}}
\DeclareRobustCommand{\Stop}{\ifmmode\text{\textsc{Stop}}\else\textsc{Stop}\fi}
\DeclareRobustCommand{\Insert}{\ifmmode\text{\textsc{Insert}}\else\textsc{Insert}\fi}

\DeclareRobustCommand{\Move}{\ifmmode\text{\textsc{Move}}\else\textsc{Move}\fi}
\DeclareRobustCommand{\Delete}{\ifmmode\text{\textsc{Delete}}\else\textsc{Delete}\fi}

\DeclareRobustCommand{\Replace}{\ifmmode\text{\textsc{Replace}}\else\textsc{Replace}\fi}
\DeclareRobustCommand{\SpanDelete}{\ifmmode\text{\textsc{SpanDelete}}\else\textsc{SpanDelete}\fi}
\DeclareRobustCommand{\Swap}{\ifmmode\text{\textsc{Swap}}\else\textsc{Swap}\fi}
\DeclareRobustCommand{\SpanMove}{\ifmmode\text{\textsc{SpanMove}}\else\textsc{SpanMove}\fi}
\DeclareRobustCommand{\SpanCopy}{\ifmmode\text{\textsc{SpanCopy}}\else\textsc{SpanCopy}\fi}

\newcommand{\Mset}{\mathcal{M}}

\newcommand{\G}{\mathrm{G}} %
\newcommand{\M}{\mathrm{M}} %

\definecolor{TokGreen}{HTML}{1B7F3A}
\definecolor{TokRed}{HTML}{B3261E}
\definecolor{FigText}{HTML}{222222}
\definecolor{FigTyrianPurple}{HTML}{66023C}
\definecolor{FigBlue}{HTML}{3B82F6}
\definecolor{FigGray}{HTML}{E5E7EB}
\definecolor{FigDarkGray}{HTML}{6B7280}
\definecolor{FigLightBlue}{HTML}{DBEAFE}
\definecolor{FigLightOrange}{HTML}{FEF3C7}
\definecolor{FigRed}{HTML}{B91C1C}

\newcommand{\TokG}[1]{\begingroup\color{TokGreen}\ifmmode #1\else\texttt{#1}\fi\endgroup}
\newcommand{\TokR}[1]{\begingroup\color{TokRed}\ifmmode #1\else\texttt{#1}\fi\endgroup}

\tikzset{
  diagramBox/.style={draw, thick, rounded corners=2pt, align=center, inner sep=6pt, minimum height=10mm, text width=35mm},
  diagramSmall/.style={draw, thick, rounded corners=2pt, align=center, inner sep=4pt, minimum height=8mm, text width=28mm},
  diagramArrow/.style={-Latex, thick},
  diagramDashed/.style={draw, thick, dashed, rounded corners=2pt},
  diagramTitle/.style={font=\bfseries},
  stepBox/.style={draw, thick, rounded corners=2pt, align=left, inner sep=5pt, text width=115mm},
  decision/.style={diamond, draw, thick, aspect=2, align=center, inner sep=2pt, text width=25mm}
}

\title{Reviser: Writing and Editing Text with\texorpdfstring{\\}{ } Autoregressive Cursor Actions}
\author{Sean Diab \\
  Independent Researcher \\
  \texttt{diabsean2005@gmail.com}}

\hypersetup{pdftitle={Reviser: Writing and Editing Text with Autoregressive Cursor Actions}, pdfauthor={Sean Diab}}
\begin{document}
\maketitle
\suppressfloats[t]

\begin{abstract}
Language models write text once, left to right, and models that write in other orders, such as diffusion LMs, pay for it with many full-sequence passes. We introduce \textbf{Reviser}, a decoder-only Transformer that writes and edits text with cursor actions (insert, move, delete, stop) that a deterministic executor applies to a canvas. Reviser is autoregressive over its own actions, so a step costs the same as an autoregressive (AR) step. We train Reviser to continue text, including a self-correction recipe that types a wrong token and deletes it, and Reviser-Edit to revise a draft, and compare each with a same-size AR model trained on the same data, using three LLM judges on continuations cut to the same length. Reviser-Edit beats its matched AR rewriter on CoEdIT SARI (+7.2 at 100M, +4.5 at 300M; two seeds agree) with a third of the decoding steps and latency, though the AR rewriter corrects grammar better. In continuation, Reviser beats diffusion LMs of similar size (62--74\% win rate) at a fraction of their compute; two of three judges prefer its AR twin, and one slightly prefers Reviser at 100M. Reviser fills gaps it left and backspaces a just-typed repeated token with probability 0.85--0.87.\footnote{Code: \url{https://github.com/Sean-Diab/Reviser}. Checkpoints: \url{https://huggingface.co/sean-diab/reviser-checkpoints}. Step-by-step replays of example trajectories: \url{https://sean-diab.github.io/reviser_trajectories.html}.}
\end{abstract}

\section{Introduction}
\label{sec:intro}
\begin{figure*}[t]
\centering
\begin{minipage}[c]{0.49\textwidth}
\centering
\begin{tikzpicture}
\begin{axis}[
  scale only axis, width=4.5cm, height=3.45cm,
  xmode=log, xmin=0.75, xmax=7000,
  ymin=0.35, ymax=11.65, y dir=reverse,
  ytick={1,2,3,4,5,6,7,8,9,10,11},
  yticklabels={\textbf{Reviser (ours)},SDLM-D4,One-shot NAT,InsT,LevT,Mask-Predict,Dream-7B,LLaDA-8B,Diffusion-LM,SEDD,MDLM},
  xtick={1,10,100,1000}, xticklabels={1$\times$,10$\times$,100$\times$,1000$\times$},
  minor xtick={2,3,4,5,6,7,8,9,20,30,40,50,60,70,80,90,200,300,400,500,600,700,800,900,2000,3000,4000,5000,6000,7000},
  xlabel={Inference FLOPs relative to AR (log scale)},
  tick label style={font=\footnotesize, text=FigText}, label style={font=\footnotesize, text=FigText},
  xlabel style={yshift=2pt},
  xmajorgrids, grid style={FigGray}, major tick length=2pt, minor tick length=1.2pt,
  axis line style={FigDarkGray}, tick style={FigDarkGray}, ytick style={draw=none},
  unbounded coords=jump, clip=false,
  legend style={font=\footnotesize, text=FigText, draw=FigDarkGray, fill=white, at={(0.985,0.985)}, anchor=north east, inner sep=1pt, row sep=-2pt},
  legend cell align=left,
]
\addplot[FigBlue!30, line width=1.5pt, forget plot] coordinates {
 (1.85,2) (2.26,2) (nan,nan) (2.02,4) (65.01,4) (nan,nan) (4.18,5) (6.34,5) (nan,nan)
 (6.09,6) (36.18,6) (nan,nan) (9.67,7) (138.23,7) (nan,nan) (61.91,8) (149.99,8) (nan,nan)
 (201.88,9) (2018.80,9) (nan,nan) (32.30,10) (2085.6,10) (nan,nan) (121.32,11) (1011.25,11)};
\addplot[only marks, mark=|, mark size=3.3pt, mark options={line width=1.1pt, draw=FigBlue!70!black}] coordinates {
 (2.0,2) (1.00,3) (2.02,4) (65.01,4) (6.34,5) (6.09,6) (29.01,6) (9.67,7) (123.34,7) (61.91,8) (123.79,8)
 (201.88,9) (2018.80,9) (32.30,10) (2067.25,10) (122.34,11) (1009.40,11)};
\addlegendentry{Analytic}
\addplot[only marks, mark=*, mark size=1.75pt, mark options={fill=FigBlue, draw=white, line width=0.4pt}] coordinates {
 (1.85,2) (2.26,2) (4.18,5) (4.20,5) (7.05,6) (36.18,6) (10.80,7) (138.23,7) (75.00,8) (149.99,8)
 (33.6,10) (2085.6,10) (121.32,11) (1011.25,11)};
\addlegendentry{Measured}
\addplot[FigTyrianPurple!30, line width=1.5pt, forget plot] coordinates {(1.25,1) (1.50,1)};
\addplot[only marks, mark=|, mark size=3.3pt, mark options={line width=1.1pt, draw=FigTyrianPurple!70!black}, forget plot] coordinates {(1.25,1) (1.50,1)};
\addplot[only marks, mark=*, mark size=1.75pt, mark options={fill=FigTyrianPurple, draw=white, line width=0.4pt}, forget plot] coordinates {(1.31,1) (1.47,1)};
\draw[FigDarkGray, dashed, thin] (axis cs:1,0.35) -- (axis cs:1,11.65);
\node[font=\footnotesize, text=FigText, anchor=south] at (axis cs:1,0.35) {AR};
\end{axis}
\end{tikzpicture}
\end{minipage}\hfill
\begin{minipage}[c]{0.497\textwidth}
\centering
\footnotesize
\setlength{\tabcolsep}{3pt}
\maxwidthbox{\begin{tabular}{@{}llrr@{}}
\toprule
Method & Decoding & Analytic & Measured \\
\midrule
AR & $n$ steps & 1 & 1 \\
\textbf{Reviser (ours)} & $n$ steps + moves & 1.25--1.50 & 1.31--1.47$^{c}$ \\
SDLM-D4 & blocks of 4 & 2.0 & 1.85--2.26$^{g}$ \\
One-shot NAT & 1 pass & 1.00 & -- \\
InsT & $\log_2 n$ or $n$ rounds & 2.02--65.0 & -- \\
LevT & 3.3 iter.\ (avg.) & 6.34 & 4.18--4.20$^{w}$ \\
Mask-Predict & 10 iter., 1--5 lengths & 6.09--29.0 & 7.05--36.2$^{w}$ \\
Dream-7B & $T{=}$10--128 & 9.67--123 & 10.8--138$^{g}$ \\
LLaDA-8B & $T{=}$64--128 & 61.9--124 & 75.0--150$^{g}$ \\
Diffusion-LM & $T{=}$200--2000 & 202--2019 & -- \\
SEDD & $T{=}$32--2048 & 32.3--2067 & 33.6--2086$^{o}$ \\
MDLM & $T{=}$1000, $\pm$cache & 122--1009 & 121--1011$^{o}$ \\
\bottomrule
\end{tabular}}
\end{minipage}
\caption{Inference FLOPs relative to an AR Transformer of the same size. \emph{Analytic}: one shared convention, at the released decoders' settings (\Cref{app:prior_art_efficiency_flops}). \emph{Measured}: counted while trained checkpoints decode at batch 1 with their released decoders, against a length-matched AR model on the same task (\Cref{app:measured_flops}): $^{c}$C4, released 100M Reviser; $^{w}$WMT14 En--De; $^{o}$OpenWebText; $^{g}$GSM8K, Qwen2.5 AR. $T$: denoising steps. FLOPs are not time: at batch 1, Dream and LLaDA ($T{=}128$) take 7--8$\times$ AR's time, Reviser 1.2--1.7$\times$ (\Cref{sec:results_speed}).}
\label{fig:efficiency}
\end{figure*}
Autoregressive (AR) Transformers write text in its final left-to-right order: once a token is emitted, a mistake can only be followed by more text \citep{vaswani2017attention,brown2020language}. Human writers instead move back, insert, and delete. Methods that relax left-to-right order---insertion and Levenshtein Transformers, iterative refinement, and diffusion LMs---usually pay for it with many full-sequence passes \citep{stern2019insertion,gu2019levenshtein,ghazvininejad2019maskpredict,lou2024sedd,sahoo2024mdlm}.

\textbf{Reviser} keeps the AR interface but changes what is predicted: a stream of cursor actions on a mutable canvas (\Insert{} a token, \Move{} the cursor by $\pm2^k$, \Delete{} the token left of the cursor, \Stop). The Transformer reads only its action history, and a deterministic executor replays the actions to obtain the text. Each step is one next-token prediction, so the cost of non-monotonic writing is the number of extra actions, not extra passes (\Cref{sec:model}). The same model can start from an existing draft and edit it, which makes revision, not only non-monotonic insertion, testable.

We evaluate Reviser in two settings. \emph{Continuation:} we train Reviser on three kinds of synthetic writing orders and compare each model with an AR model of the same size trained on the same text, and with public AR and diffusion LMs (\Cref{sec:training,sec:results_cont}). \emph{Editing:} we train Reviser-Edit and a matched AR rewriter on the same 1B-token edit mix and compare them on CoEdIT, IteraTeR and JFLEG (\Cref{sec:results_edit}).

Our main findings:
\begin{itemize}[leftmargin=*,itemsep=0.1em,topsep=0.2em]
\item \textbf{Editing with a cursor beats rewriting.} Reviser-Edit beats a matched AR rewriter on CoEdIT SARI at both sizes and two training seeds with a third of its decoding steps and latency, and is preferred by an LLM judge at 100M; the AR rewriter is better at grammar correction and at leaving correct text alone.
\item \textbf{Non-monotonic writing at near-AR cost.} A Reviser step costs the same as an AR step, and Reviser needs 1.2--1.5 steps per kept token. Measured on trained models, it uses 1.3--1.5$\times$ the FLOPs of AR, while refinement and diffusion LMs at their default settings use 4--150$\times$ (\Cref{fig:efficiency}). With both texts cut to the same length, Reviser beats diffusion LMs of similar size (62--74\%); against its AR twin, Codex and Claude prefer the AR model, and Skywork slightly prefers Reviser at 100M (52--54\%).
\item \textbf{The cursor is used.} Reviser models move backward and fill gaps they left. Trained with self-correction, they backspace a just-typed repeated token with probability 0.85--0.87 and a correct one with 0.01--0.04, and in free writing they replace improbable tokens with likelier ones; plausible substitutes are rarely caught.
\end{itemize}

\paragraph{Contributions.}
(i)~Reviser, a history-only cursor-action LM with \Insert, \Move, \Delete{} and \Stop, and a prompt lock (\Cref{sec:model}).
(ii)~Three synthetic writing orders for continuation, a self-correction recipe for \Delete, RL fine-tuning, and diff trajectories for editing (\Cref{sec:training}).
(iii)~A matched evaluation: AR twins trained on the same data, three judges at matched length in both orders, ablations of move rate, writing order and deletion, and measured FLOPs and latency (\Cref{sec:setup_eval,sec:results_cont,sec:results_ablate,sec:results_speed}).

\section{Related Work}
\label{sec:related}
\paragraph{Non-left-to-right generation.} One-shot and iterative NAR models \citep{gu2018nonautoregressive,lee2018deterministic,ghazvininejad2019maskpredict}, insertion-based models \citep{stern2019insertion,gu2019insertion,welleck2019nonmonotonic,shen2020blank}, the Levenshtein Transformer \citep{gu2019levenshtein}, Insertion Language Models, which insert one token at any position per step \citep{patel2025insertion}, latent-variable NAR \citep{ma2019flowseq} and diffusion LMs \citep{li2022diffusionlm,austin2021structured,lou2024sedd,sahoo2024mdlm,ye2025dream} decouple generation order from text order, usually with repeated full-sequence passes (\Cref{fig:efficiency} compares their compute). XLNet permutes the factorization order \citep{yang2019xlnet}, and infilling models write a missing span given both sides \citep{donahue2020ilm,bavarian2022fim}. Reviser instead makes the action history the autoregressive object, one cursor-relative action per step. SequenceMatch adds a backspace action to an AR LM and trains it by imitation \citep{cundy2024sequencematch}, and pretraining on erroneous steps followed by their corrections teaches LMs to fix their mistakes \citep{ye2025physics22}; our self-correction recipe is a supervised, token-level variant of these ideas inside a cursor model.

\paragraph{Text editing.} Edit taggers and span-edit models predict edits over an input \citep{malmi2019lasertagger,mallinson2020felix,omelianchuk2020gector,stahlberg2020seq2edits}, and LLMs can be fine-tuned to output edit spans instead of the full target, which shortens the output at comparable quality \citep{kaneko2023reducing}. Instruction-based editors such as PEER and CoEdIT rewrite the text with a sequence-to-sequence model \citep{schick2023peer,raheja2023coedit}, and IteraTeR studies human iterative revision \citep{du2022iterater}. LLMs can also refine their own drafts by re-generation \citep{madaan2023selfrefine}. Reviser-Edit applies cursor edits directly to the draft, so unchanged text costs no output tokens.

\paragraph{LLM judges.} Pairwise LLM judges show position and length biases \citep{zheng2023judging,dubois2024lengthcontrolled}. We judge in both orders and cut both texts to the same length.

\section{Reviser}
\label{sec:model}
\paragraph{State and actions.} At step $t$ the state is $s_t=(C_t,u_t,H_t)$: a canvas $C_t=(c_1,\dots,c_{\ell_t})$, a cursor boundary $u_t\in\{0,\dots,\ell_t\}$, and the action history $H_t$. The action vocabulary is $V_a=V_c\cup\Mset\cup\{\Delete,\Stop\}$, where each $x\in V_c$ (the GPT-2 vocabulary) is $\Insert(x)$ and $\Mset=\{\pm2^k:k=0,\dots,9\}$ are relative moves; a longer move is a sequence of moves.

\paragraph{Executor.} $\Insert(x)$ places $x$ at the cursor and advances it ($u_{t+1}=u_t+1$). $\Move(\Delta)$ sets $u_{t+1}=u_t+\Delta$. $\Delete$ is a backspace: it removes $c_{u_t}$ and sets $u_{t+1}=u_t-1$. $\Stop$ returns $C_t$. Deleting $k$ tokens takes $k$ actions; there is no replace or span operator (\Cref{sec:optional_ops} lists such extensions).

\paragraph{Validity mask and prompt lock.} For continuation, the $m$-token prompt is written by $m$ \Insert{} actions, leaving the cursor at $m$. A mask computed from $(C_t,u_t)$ sets invalid actions to $-\infty$: $\Move(\Delta)$ is valid only if $m\le u_t+\Delta\le\ell_t$, and \Delete{} only if $u_t>m$, so the prompt cannot change. Decoding and RL always apply the mask; in training, our continued models put no loss on the prompt and the \Delete{} models also apply the mask (\Cref{sec:repro_hparams}). The earlier version's decoder had no lock; \Cref{tab:main,tab:lmcompare} are re-decoded with it except the $\dagger$ cells and the earlier full-length column, while the ablations (\Cref{tab:ablate}) and cursor probes keep the earlier mask, which lets a model edit the prompt.

\paragraph{Model.} A standard causal Transformer (RoPE, tied embeddings) reads the action history and predicts $p(a_t\mid H_t)=\mathrm{softmax}(W h_{t-1}+m_t)$ with mask $m_t$. The trunk never reads the canvas: the canvas is implicit in the history and is replayed on the CPU only to build the mask. A step therefore costs one KV-cached decoder call, like an AR step; producing $n$ tokens takes $n$ steps plus one step per move, per delete and per deleted insert. \Cref{fig:reviser_step} shows one step, \Cref{alg:reviser_decode} the loop.

\begin{figure*}[t]
\centering
\resizebox{\textwidth}{!}{%
\begin{tikzpicture}[
font=\normalsize,
>=Latex,
box/.style={draw=FigDarkGray, rounded corners=2pt, thick, fill=white, inner sep=6pt, align=center},
mini/.style={draw=FigDarkGray, rounded corners=2pt, thick, fill=white, inner sep=4pt, align=center},
arrow/.style={->, thick, draw=FigDarkGray},
every node/.style={text=FigText}
]
\node[box, text width=3.6cm, align=center] (state) {\textbf{Current state at step $t$}};
\node[box, right=1.2cm of state, text width=3.2cm, align=center] (model) {\textbf{Decoder-only Transformer}\\[3pt]autoregressive over\\edit-history actions $H_t$};
\node[mini, below=0.55cm of model, text width=3.2cm] (mask) {\textbf{Validity mask} from \textbf{$(C_t,u_t)$}\\invalid moves/deletes, prompt lock};
\node[box, right=1.2cm of model, text width=3.3cm] (action) {\textbf{Next-action}\\\textbf{distribution}\\[2pt]{\normalsize\textcolor{FigTyrianPurple}{\textsc{Insert}(\texttt{language})}}\\{\normalsize\textcolor{FigBlue}{\textsc{Move}(+2)}}\\{\normalsize\textcolor{FigRed}{\textsc{Delete}}}\quad{\normalsize\textcolor{FigRed}{\textsc{Stop}}}};
\node[mini, fill=white, draw=FigDarkGray, text width=3.6cm] (chosen) at (action |- mask) {\textbf{Chosen action $a_t$}\\{\normalsize\textcolor{FigTyrianPurple}{\textsc{Insert}(\texttt{language})}}};
\node[box, right=1.2cm of action, text width=3.6cm] (exec) {\textbf{Executor}\\[2pt]Apply $a_t$ to update\\canvas, cursor, and history};
\node[box, text width=4.0cm] (updated) at (exec |- mask) {\textbf{Updated state at step $t+1$}};
\coordinate (hcenter) at ($(state.west)!0.5!(exec.east)$);
\node[draw=FigDarkGray, rounded corners=2pt, thick, fill=white, inner sep=5pt, anchor=center, text width=15cm, align=center] (input_state) at ($(hcenter) + (0, 2.3cm)$) {Canvas: \texttt{The | model can revise earlier text.}\\Cursor: $u_t$\quad History: ${H_t=(a_1,\dots,a_{t-1})}$};
\node[draw=FigDarkGray, rounded corners=2pt, thick, fill=white, inner sep=5pt, anchor=center, text width=16.9cm, align=center] (output_state) at ($(hcenter |- updated.south) + (0, -1.4cm)$) {Canvas: \texttt{The language | model can revise earlier text.}\\Cursor: $u_{t+1}$\quad History: ${H_{t+1}=(a_1,\dots,a_t)}$};
\draw[arrow] ($(input_state.south -| state.north) + (0.7cm, 0)$) -- (state.north);
\draw[arrow] (state) -- (model);
\draw[arrow] (model) -- (action);
\draw[arrow] ($(state.south east) + (-1pt, 1pt)$) -- ($(mask.north west) + (1pt, -1pt)$);
\draw[arrow] ($(mask.south east)!0.75!(mask.north east)$) -- ($(action.south west) + (1pt, 1pt)$);
\draw[arrow] (action.south) -- (chosen.north);
\draw[arrow] ($(chosen.north east) + (-1pt, -1pt)$) -- ($(exec.south west) + (1pt, 1pt)$);
\draw[arrow] (exec) -- (updated);
\draw[arrow] (updated.south) -- ($(updated.south |- output_state.north) + (-0.7cm, 0)$);
\end{tikzpicture}%
}%
\caption{One Reviser decoding step. The model reads only the action history and predicts one action (\Insert, \Move, \Delete{} or \Stop); a mask built from the replayed canvas removes invalid actions, including any edit of the prompt; a deterministic executor applies the action.}
\label{fig:reviser_step}
\end{figure*}

\section{Training Trajectories}
\label{sec:training}
Reviser is trained by teacher forcing on action trajectories that end in the target text; the loss covers the actions after the prompt. The trajectories are synthetic: they define \emph{an} order in which the text could be written, not recorded human keystrokes.

\paragraph{Random obfuscation--restoration.} The released models of the earlier version of this work start from the target with a random cursor and repeatedly delete the token left of the cursor (probability 0.8) or make a random valid move (0.2) until the canvas is empty; inverting this walk gives a restoration trajectory of inserts and moves (\Cref{alg:obf_restore} and \Cref{tab:trajectory}). Its moves are random.

\paragraph{Run-and-jump order.} A forward generator writes the text in runs: with probability $\lambda=0.82$ it types the next token after the cursor; otherwise it jumps to an unwritten gap, chosen with weight $(1+d)^{-1.5}$ by distance $d$, and may skip ahead to leave a hole that a later backward jump fills (probability 0.3; a far start in the unwritten tail with 0.9). It inserts every token exactly once, with 0.33 moves per token, 69\% of them backward. Its dials were chosen to resemble the restoration statistics, not human writing; \Cref{sec:results_ablate} varies them (\Cref{app:trajdata}).

\paragraph{Self-correction with \Delete.} On top of the run-and-jump order, each continuation token gets, with probability 0.05, a wrong token $w$ drawn from the same text within $\pm4$ positions. The writer types $w$, immediately deletes it, and types the right token. The label of $w$ is masked, so the model learns to delete wrong tokens without learning to write them; \Delete{} and the retyped token are trained. This gives 0.046 deletes per token.

\paragraph{RL fine-tuning.} We fine-tune the self-correction model for 200 steps with a group-relative policy gradient (GRPO-style; \citealp{shao2024deepseekmath}): 16 prompts $\times$ 8 samples per step, advantages normalized within each group, one on-policy update, and an exact KL penalty (0.2) to the starting model; no clipping and no value head. The reward is
$r=-\mathrm{NLL}_{\text{Qwen}}-8\,\mathrm{rep}_4-\max(0,1-n/150)-2\max(0,f-\mu)$,
where $\mathrm{NLL}_{\text{Qwen}}$ is the mean token NLL of the continuation under Qwen2.5-3B-Instruct \citep{qwen2024qwen25}, $\mathrm{rep}_4$ the share of repeated 4-grams, $n$ the number of new tokens, $\mu$ the moves per new token and $f$ its value at the first step (so RL cannot simply stop moving). The reward is a likelihood, so our perplexity metrics---especially Dream-7B, which is built on Qwen2.5---are \emph{not} independent of it, and MAUVE (BERT likelihood and length features) only partly; the LLM judges are. Details: \Cref{sec:rl}.

\paragraph{Editing trajectories (Reviser-Edit).} For an edit pair (source $\to$ target) with an optional instruction, the canvas starts as the source with the cursor at its end. We compute the diff hunks and edit them right to left: move to the end of a hunk, \Delete{} its old tokens, insert the new ones; then \Stop. An unchanged input is a single \Stop. The \textbf{matched AR rewriter} sees the instruction and source as a prompt and writes the target left to right. Both models are trained on the same examples in the same order for the same number of steps (187,797 steps of 64 episodes, about 1B tokens of source plus target text over 3 passes); they differ only in output layout and initialization (Reviser-Edit from the released Reviser, the AR rewriter from our AR baseline of the same size). Because it copies unchanged text for free, Reviser-Edit receives fewer supervised labels (247M vs.\ 531M). The mix (\Cref{app:editdata}) contains CoEdIT and IteraTeR train splits, C4\_200M grammar errors \citep{stahlberg2021c4200m}, restoration, no-change and synthetic repair examples, and edit drafts that Qwen2.5-32B-Instruct and Claude Haiku 5.5 wrote from clean paragraphs; evaluation sets were removed.

\section{Experimental Setup}
\label{sec:setup_eval}
\paragraph{Models.} All models have 100M ($L{=}24,d{=}512$) or 300M ($L{=}26,d{=}896$) parameters. The AR baselines share the Reviser architecture and were trained on FineWeb \citep{penedo2024fineweb} text; the released Reviser models were trained on restoration trajectories of FineWeb text.
Their training budgets are not matched (\Cref{sec:repro_hparams}). Starting from each released model, we continue training for 608.7M tokens on the run-and-jump order with or without self-correction, using the same text and seed (\Cref{tab:main}), and apply RL to the self-correction model (three RL seeds at 100M, one at 300M). The AR baselines get no extra training, which, if anything, favors the continued Reviser models.

\paragraph{Continuation.} We use 3,000 C4 validation prompts \citep{raffel2020t5} with a 35-token prompt (3 seeds $\times$ 1,000), sample with temperature 0.9 and top-$k$ 50 for at most 220 new actions, and decode every Reviser model with the prompt lock. We report evalPPL of the continuation under GPT-2 \citep{radford2019gpt2} and Dream-7B \citep{ye2025dream}, MAUVE \citep{pillutla2021mauve} on three features per text (mean and spread of its BERT \citep{devlin2019bert} pseudo-log-likelihoods, and its length), and pairwise judgments. \textbf{Judges.} Skywork-Critic-Llama-3.1-8B \citep{skywork2024critic} judges all 3,000 pairs two ways: the original protocol (full texts, one random order) and \emph{matched length} (both continuations cut to the shorter one's length, both orders, a split counts as a tie). Codex (gpt-6-astra) and Claude (Opus 5.5) judge 300 matched-length pairs in both orders, blind to the systems, with an instruction that asks for coherence with the prompt, fluency and absence of repetition and not to prefer longer texts (\Cref{app:judges}). Scores are the Reviser model's share (tie $=0.5$) with 95\% bootstrap CIs.

\paragraph{Editing.} CoEdIT validation \citep{raheja2023coedit} (1,712 items: grammar, paraphrase, neutralize), IteraTeR human test sentences \citep{du2022iterater} (364), JFLEG test \citep{napoles2017jfleg} (747) and 1,000 synthetic repairs. We report SARI \citep{xu2016sari}, exact match, ERRANT F$_{0.5}$ \citep{bryant2017errant} on CoEdIT grammar items and IteraTeR fluency items (our setup, not the BEA-2019 protocol), JFLEG GLEU \citep{napoles2015gleu}, the share of inputs left unchanged when the reference itself is given as the input, and a Codex judge (300 items, both orders). Decoding is greedy over action types: we take the type (\Insert, \Move, \Delete{} or \Stop) with the largest total probability, then its most likely token. References: copying the input, CoEdIT-large, and zero-shot Qwen2.5-0.5B/1.5B-Instruct and SmolLM2-360M-Instruct \citep{benallal2025smollm2}.

\section{Results: Writing a Continuation}
\label{sec:results_cont}

\begin{table*}[t]
\centering
\small
\setlength{\tabcolsep}{3pt}
\resizebox{\textwidth}{!}{%
\begin{tabular}{l rrr rrr rrr rrr}
\toprule
 & \multicolumn{3}{c}{Behavior} & \multicolumn{3}{c}{Quality} & \multicolumn{3}{c}{AR cut to the same length} & \multicolumn{3}{c}{Judges vs.\ AR} \\
\cmidrule(lr){2-4}\cmidrule(lr){5-7}\cmidrule(lr){8-10}\cmidrule(lr){11-13}
Model & Len & Mv/tok & Del/tok & GPT-2$\downarrow$ & Dream$\downarrow$ & MAUVE$\uparrow$ & GPT-2$\downarrow$ & Dream$\downarrow$ & MAUVE$\uparrow$ & Skywork & Codex & Claude \\
\midrule
AR 100M & 220.0 & 0 & 0 & 13.40 & 13.85 & 0.935 & -- & -- & -- & -- & -- & -- \\
Reviser, released & 67.8 & 0.465 & 0 & 32.33 & 25.68 & 0.811 & 17.80 & 18.85 & 0.921 & 53.9 & 0.200 & 0.158 \\
\quad + run-and-jump & 101.1 & 0.188 & 0 & 23.19 & 19.97 & 0.931 & 16.06 & 16.91 & 0.940 & 52.3 & 0.242 & 0.175 \\
\quad + R\&J + \Delete & 113.6 & 0.152 & 0.033 & 22.22 & 18.88 & 0.927 & 15.43 & 16.31 & 0.936 & 53.0 & 0.215 & 0.192 \\
\quad \quad + RL (3 seeds) & 142.1 & 0.140 & 0.029 & 18.38$^\ast$ & 15.56$^\ast$ & 0.931 & 14.50 & 15.38 & 0.900 & 53.3 & 0.286 & 0.268 \\
\midrule
AR 300M & 220.0 & 0 & 0 & 12.63 & 12.12 & 0.844 & -- & -- & -- & -- & -- & -- \\
Reviser, released & 94.7 & 0.381 & 0 & 17.96 & 16.03 & 0.901 & 14.59 & 14.61 & 0.853 & 48.8 & 0.293$^\dagger$ & 0.233$^\dagger$ \\
\quad + R\&J + \Delete & 110.4 & 0.147 & 0.029 & 17.00 & 15.35 & 0.948 & 14.32 & 14.25 & 0.851 & 44.5 & 0.273 & 0.193 \\
\quad \quad + RL (seed 3) & 141.2 & 0.146 & 0.026 & 14.91$^\ast$ & 13.49$^\ast$ & 0.900 & 13.59 & 13.59 & 0.844 & 45.2 & 0.308 & 0.287 \\
\bottomrule
\end{tabular}}
\caption{C4 continuation (3,000 prompts, prompt-locked decodes, T 0.9, top-$k$ 50, $\le$220 actions). Released: the random-order model of the earlier version; indented rows continue it for 608M tokens (R\&J: run-and-jump). Len: mean new tokens (share ending with \Stop: 99.5, 94.3, 87.9, 74.9\% at 100M; 98.3, 90.6, 79.1\% at 300M); Mv/Del per output token. evalPPL under GPT-2 base and Dream-7B (8 MC samples). MAUVE: standard deviation 0.004--0.008 over 10 k-means seeds. \emph{AR cut to the same length}: the AR model's text for each prompt cut to the length of that Reviser model's text (GPT-2 tokens), scored the same way. Judges: the Reviser model's share against the AR model of the same size, both texts cut to the same length and judged in both orders (tie $=0.5$); Codex/Claude on 300 items; full-text Skywork scores: \Cref{app:fulllength}. 95\% CIs: Skywork $\pm$1.6, Codex/Claude $\pm$0.04--0.05. RL seeds 1/2/3: Skywork 52.1/53.4/54.4, Codex 0.250/0.268/0.340, Claude 0.230/0.261/0.313. $^\ast$The RL reward includes a Qwen2.5-3B likelihood, so these perplexities are not independent of training. $^\dagger$Stored outputs of the earlier version (repeated prompt removed) vs.\ the stored AR 300M outputs.}
\label{tab:main}
\end{table*}

\paragraph{Reviser against its AR twin.} \Cref{tab:main} compares every Reviser model with the AR model of the same size. Reviser stops earlier (68--142 vs.\ 220 tokens) and judges favor longer texts \citep{dubois2024lengthcontrolled}, so our main comparison cuts both texts to the same length and judges both orders. Skywork then slightly prefers Reviser to its AR twin at 100M (52--54\%; 95\% CI $\pm$1.6) and rates it a little lower at 300M (44.5--48.8\%), while Codex and Claude prefer the AR model (Reviser's share 0.16--0.31). On full texts, the protocol of the earlier version of this work (\Cref{app:fulllength}), Skywork prefers the released Reviser more (62.5\% at 100M, 53.2\% at 300M); part of that lead comes from length. With each AR text cut to the length of the paired Reviser text, AR keeps a lower GPT-2 evalPPL (14.5 vs.\ 18.4 after RL at 100M), but after RL Reviser matches it under Dream-7B (15.6 vs.\ 15.4 at 100M, 13.5 vs.\ 13.6 at 300M; partly the reward) and has a higher MAUVE (0.931 vs.\ 0.900 at 100M; 0.90--0.95 vs.\ 0.84--0.85 for every 300M model), whose features partly overlap the reward.

\paragraph{Continued training improves Reviser.} Continuing the released 100M model on the run-and-jump order lowers GPT-2 evalPPL from 32.3 to 23.2 and raises MAUVE from 0.81 to 0.93, close to AR's 0.935, with longer outputs (length is a MAUVE feature) and fewer moves. Self-correction lowers evalPPL further (22.2), and RL to 18.4 (AR: 13.4); the RL gain in perplexity is partly the reward itself (\Cref{sec:training}), and the judges move less than the perplexities. At 300M, with a 10$\times$ lower learning rate, the loss barely changed (3.85 $\to$ 3.82); Skywork rates them below the released one, Codex and Claude about the same.

\paragraph{Comparison with other LMs.} %
\Cref{tab:lmcompare} re-runs the earlier comparisons at matched length. Reviser beats the diffusion LMs (62--74\%; their texts have GPT-2 evalPPL 99--145), which decode with 128 passes over the whole sequence, as well as Cerebras-GPT (66--69\%) and Pythia 160M (whose stored texts are degenerate), ties GPT-2 117M (52\%) and is just behind OPT, Pythia 410M and GPT-2 Medium (45--47\%). Codex and Claude agree on the diffusion LMs (Reviser's share 0.64--0.97 and 0.55--0.96; the closest pair is the released 100M against the 3.5$\times$ larger SEDD Absorb), and Codex prefers GPT-2 (0.24--0.35).

\begin{table}[t]
\begin{minipage}[t]{0.485\textwidth}
\vspace{0pt}
\centering
\footnotesize
\setlength{\tabcolsep}{4pt}
\maxwidthbox{%
\begin{tabular}{@{}l rrrr@{}}
\toprule
 & & \multicolumn{3}{c}{Reviser's win rate (\%)} \\
\cmidrule(lr){3-5}
 & & Full & \multicolumn{2}{c}{Matched length} \\
\cmidrule(lr){4-5}
Opponent (new tokens) & PPL & (earlier) & released & +RL$^\ddagger$ \\
\midrule
\multicolumn{5}{@{}l}{\emph{vs.\ Reviser 100M}} \\
SEDD Small 169M (141) & 133.9 & 85.9$^\ast$ & 68.9 & 63.3 \\
MDLM 170M (139) & 144.7 & 78.3 & 73.9 & 69.0 \\
Cerebras-GPT 111M (215) & 16.3 & 72.1 & 66.5 & 66.8 \\
Pythia 160M (220) & 78.5 & 68.8 & 63.9 & 62.7 \\
GPT-2 117M (202) & 10.5$^\S$ & 58.6 & 52.1 & 52.3 \\
OPT 125M (199) & 13.2 & 53.9 & 46.8 & 47.2 \\
\midrule
\multicolumn{5}{@{}l}{\emph{vs.\ Reviser 300M}} \\
SEDD Absorb 353M (135) & 98.7 & 68.5 & 66.8 & 62.3 \\
Cerebras-GPT 256M (219) & 13.2 & 71.3 & 69.0 & 67.7 \\
Pythia 410M (220) & 14.1 & 51.3 & 45.9 & 44.9 \\
GPT-2 Medium 345M (202) & 13.1 & 50.4 & 46.7 & 45.8 \\
OPT 350M (195) & 12.5 & 49.9 & 46.7 & 45.5 \\
\bottomrule
\end{tabular}}
\caption{Reviser's Skywork win rate (\%) against public LMs \citep{lou2024sedd,sahoo2024mdlm,cerebras2023,biderman2023pythia,radford2019gpt2,zhang2022opt}. Full: the earlier full-length protocol (one order). Matched: both texts cut to the same length, both orders ($\pm$1.6); same 3,000 prompts and stored baseline texts (diffusion models: 128 steps). Released: the released Reviser; +RL: run-and-jump + \Delete{} + RL. $^\ast$Judged without the prompt in the earlier version; the same texts with the prompt give 73.1. $^\ddagger$Mean of the three RL seeds at 100M (seeds differ by $\le$2.6), seed 3 at 300M. PPL: GPT-2 evalPPL of the opponent texts ($^\S$scored by itself).}
\label{tab:lmcompare}
\end{minipage}\hfill
\begin{minipage}[t]{0.485\textwidth}
\vspace{0pt}
\centering
\footnotesize
\setlength{\tabcolsep}{4pt}
\maxwidthbox{%
\begin{tabular}{@{}l rrrr@{}}
\toprule
100M, C4 & Random & R\&J & +Del. & +RL \\
\midrule
Actions per token & 1.51 & 1.19 & 1.23 & 1.20--1.21 \\
Move share (\%) & 32.7 & 15.7 & 12.7 & 11.4--12.0 \\
Delete share (\%) & 0 & 0 & 2.7 & 2.3--2.5 \\
Moves backward (\%) & 65.5 & 79.6 & 80.6 & 79.9--81.1 \\
Moves undone (\%) & 16.4 & 0.0 & 0.0 & 0.0 \\
Insert position & 0.41 & 0.73 & 0.76 & 0.75--0.76 \\
Inserts at end (\%) & 3.9 & 35.5 & 38.8 & 34.6--36.9 \\
Opening run (tok.) & 1 & 26 & 33 & 41$^\ddagger$ \\
Backward fill (\%) & 100 & 99.9 & 99.8 & 99.9--100 \\
\bottomrule
\end{tabular}}
\caption{Trajectory statistics of the 100M models in \Cref{tab:main}: released (Random), run-and-jump (R\&J), + \Delete{}, + RL (range over 3 seeds; $^\ddagger$seed 3). Insert position is the mean position relative to the canvas length; a backward fill is a backward move followed by an insert before the end. The released 300M model is similar to the released 100M (28\% moves, 64.5\% backward).}
\label{tab:traj}
\end{minipage}
\end{table}

\paragraph{The cursor is used.} \Cref{tab:traj} summarizes the trajectories, and \Cref{sec:appendix_ranked_examples} shows examples (step-by-step replays: \url{https://sean-diab.github.io/reviser_trajectories.html}). The released model moves often and erratically (33\% of actions; 16\% of moves are undone by the next one). Models trained on the run-and-jump order write an opening run of 26--41 tokens, then move less (11--16\% of actions), mostly backward, and still place 61--65\% of their inserts before the end of the text. Almost every delete of the self-correction models (all but 15 of 68,526) removes the token typed just before it, and 94\% are followed by a different token. The deleted tokens are improbable (GPT-2 probability below 0.01 for 68\%, against 23\% of kept tokens), and the replacement is the likelier token in 68--71\% of deletes. To test \Delete{} directly, we plant one mistake as if the model had just typed it and measure the probability that the next action is \Delete{} (\Cref{app:deletes}). The self-correction models of \Cref{tab:main} delete a repeat of the token to its left with probability 0.85--0.87, a nearby-text token like those in training with 0.51--0.53 and a random token with 0.14--0.19, but a plausible substitute sampled from GPT-2 only with 0.05, about as often as a correct one (0.04). Models trained without self-correction never delete. \Delete{} is thus a local skill: it does not make the final texts less repetitive (RL does).

\section{Ablations: Writing Order and Deletion}
\label{sec:results_ablate}
To isolate the writing order, we continue the released 100M model for 84M tokens on each of 34 trajectory variants of the same texts (same seed, same learning-rate schedule) and compare each with the run-and-jump order at the same point. \Cref{tab:ablate} shows the arms that change one factor (defined in \Cref{app:trajdata}); the others agree with them. Claude and Codex judge 100 prompts (50 C4, 50 DCLM; \citealp{li2024dclm}) at matched length in both orders against the run-and-jump model; 0.5 is a tie and the 95\% CI is about $\pm$0.07. The cursor probe uses 400 held-out states: \emph{gaps filled} is the share of holes in a partial run-and-jump draft that get at least one new token, and P(\Delete) is measured right after a random token, after a repeated token, and on clean text.

\begin{table*}[t]
\centering
\small
\setlength{\tabcolsep}{3.5pt}
\resizebox{\textwidth}{!}{%
\begin{tabular}{ll rrrrrr}
\toprule
Group & Variant (change from run-and-jump) & Mv/tok & C4 PPL$\downarrow$ & Gaps filled & P(\Delete) rand./rep./clean & Claude & Codex \\
\midrule
Reference & run-and-jump, $\lambda$ = 0.82 & 0.205 & 26.40 & 0.401 & 0 / 0 / 0 & 0.5 & 0.5 \\
\midrule
Move rate & left to right only ($\lambda$ = 1) & 0 & 16.26 & 0.144 & 0 / 0 / 0 & 0.740 & 0.755 \\
 & $\lambda$ = 0.98 & 0.016 & 18.48 & 0.293 & 0 / 0 / 0 & 0.755 & 0.790 \\
 & $\lambda$ = 0.95 & 0.042 & 19.07 & 0.368 & 0 / 0 / 0 & 0.745 & 0.725 \\
 & $\lambda$ = 0.90 & 0.089 & 22.49 & 0.377 & 0 / 0 / 0 & 0.662 & 0.650 \\
 & $\lambda$ = 0.86 & 0.145 & 25.29 & 0.397 & 0 / 0 / 0 & 0.595 & 0.625 \\
 & $\lambda$ = 0.82, data rebuilt & 0.189 & 25.82 & 0.417 & 0 / 0 / 0 & 0.525 & 0.560 \\
 & $\lambda$ = 0.70 & 0.217 & 28.66 & 0.420 & 0 / 0 / 0 & 0.505 & 0.440 \\
 & + jumps between gaps ($p_{\text{fill}}$ = 0.1) & 0.324 & 27.82 & 0.441 & 0 / 0 / 0 & 0.535 & 0.517 \\
\midrule
Order & human-fitted statistics (CoAuthor) & 0.010 & 16.99 & 0.175 & 0 / 0 / 0 & 0.778 & 0.795 \\
 & neural human order (CoAuthor) & 0.026 & 18.56 & 0.286 & 0 / 0 / 0 & 0.795 & 0.790 \\
 & random order, 12\% junk inserts & 0.347 & 32.27 & 0.383 & 0.75 / 0.03 / 0.03 & 0.468 & 0.525 \\
\midrule
\Delete{} rate & 2\% near errors & 0.195 & 25.99 & 0.434 & 0.07 / 0.71 / 0.02 & 0.540 & 0.540 \\
 & \textbf{5\% (main recipe)} & 0.166 & 25.56 & 0.430 & 0.16 / 0.82 / 0.04 & 0.625 & 0.615 \\
 & 5\%, exact rerun & 0.167 & 27.45 & 0.400 & 0.17 / 0.82 / 0.04 & 0.527 & 0.575 \\
 & 10\% & 0.157 & 27.00 & 0.401 & 0.27 / 0.89 / 0.07 & 0.585 & 0.590 \\
\Delete{} timing & fixed 1 token later & 0.225 & 29.05 & 0.396 & 0.00 / 0.01 / 0.00 & 0.555 & 0.570 \\
 & fixed 3 tokens later & 0.265 & 30.44 & 0.383 & 0.00 / 0.01 / 0.00 & 0.475 & 0.420 \\
 & fixed after the last token & 0.220 & 34.55 & 0.388 & 0.00 / 0.00 / 0.00 & 0.380 & 0.415 \\
Wrong token & unigram sample & 0.175 & 25.38 & 0.376 & 0.43 / 0.15 / 0.03 & 0.573 & 0.535 \\
 & GPT-2 sample & 0.168 & 27.67 & 0.393 & 0.09 / 0.10 / 0.05 & 0.650 & 0.635 \\
 & near, wrong insert not masked & 0.190 & 28.49 & 0.391 & 0.17 / 0.82 / 0.04 & 0.610 & 0.620 \\
 & 5\% near errors, left to right & 0 & 17.66 & 0.145 & 0.25 / 0.81 / 0.13 & 0.820 & 0.800 \\
\bottomrule
\end{tabular}}
\caption{Ablations at 100M (84M tokens from the released model, one run per variant). Mv/tok: cursor moves per output token on C4. C4 PPL: GPT-2 evalPPL on 1,000 C4 prompts. Gaps filled and P(\Delete) from the cursor probe (400 held-out states). \Delete{} rows fix the wrong token immediately and mask its label unless stated. Claude/Codex: matched-length preference over the reference on 100 prompts (50 C4, 50 DCLM), both orders ($\pm$0.07). The exact rerun of the main recipe shows the run-to-run spread ($\sim$2 PPL, $\sim$0.1 judge score). Decodes without the prompt lock (\Cref{sec:model}).}
\label{tab:ablate}
\end{table*}

\paragraph{Moves teach the cursor.} Moves are what teach the model to fill gaps: the share of gaps filled rises from 0.14 without moves to 0.42 as $\lambda$ (the chance to keep typing) falls to 0.70. Text quality goes the other way: evalPPL falls from 28.7 to 16.3 as the model moves less, and both judges prefer the less-moving models (up to 0.76/0.79 against the reference). Orders fitted to human keystrokes from CoAuthor \citep{lee2022coauthor} move rarely (0.01--0.03 moves per token) and write about as well as left to right. Self-correction works with either order: added to left-to-right writing, it keeps the backspace (P(\Delete) 0.81 after a repeated token) and has the highest judge scores in \Cref{tab:ablate} (0.82/0.80; left to right alone: 0.74/0.76, within run-to-run noise).

\paragraph{How to learn \Delete.} Fixing the wrong token immediately gives by far the strongest immediate backspace (P(\Delete) 0.82 after a repeated token vs.\ 0.04 on clean text). Models trained to fix it later still delete in free writing, but our probe looks only at the next action, so it does not test fixes at their trained delay; delayed fixes also cost quality. Wrong tokens drawn from nearby text teach the model to delete repetitions; unigram tokens teach it to delete random junk; GPT-2 samples teach little of either. Masking the loss on the wrong insert gives lower evalPPL than training on it (25.6 vs.\ 28.5). A rerun of the 5\% arm moves by 1.9 PPL and 0.04--0.10 in judge score, so only the move-rate trend and the timing effect are clearly beyond run-to-run noise.

\section{Results: Editing a Draft}
\label{sec:results_edit}

\begin{table*}[t]
\centering
\small
\setlength{\tabcolsep}{3.5pt}
\resizebox{\textwidth}{!}{%
\begin{tabular}{l rrr rr r rr r}
\toprule
 & \multicolumn{3}{c}{SARI / EM $\uparrow$} & \multicolumn{2}{c}{ERRANT F$_{0.5}\uparrow$} & JFLEG$\uparrow$ & \multicolumn{2}{c}{Reference as input kept $\uparrow$} & Codex$\uparrow$ \\
\cmidrule(lr){2-4}\cmidrule(lr){5-6}\cmidrule(lr){8-9}
System & CoEdIT & EM & IteraTeR & Grammar & Fluency & GLEU & IteraTeR & CoEdIT & vs.\ AR \\
\midrule
Copy the input & 57.9 & 0.000 & 66.7 & 0.000 & 0.000 & 40.5 & 1.00 & 1.00 & -- \\
CoEdIT-large (T5, 770M) & 68.6 & 0.211 & 58.6 & 0.434 & 0.114 & 59.4 & -- & -- & -- \\
Qwen2.5-0.5B-Instruct (0-shot) & 42.4 & 0.001 & 29.2 & 0.253 & 0.025 & 41.2 & 0.02 & 0.06 & -- \\
Qwen2.5-1.5B-Instruct (0-shot) & 38.6 & 0.002 & 22.9 & 0.216 & 0.020 & 48.0 & 0.00 & 0.01 & -- \\
SmolLM2-360M-Instruct (0-shot) & 45.9 & 0.004 & 44.2 & 0.282 & 0.046 & 53.7 & 0.15 & 0.36 & -- \\
\midrule
AR rewriter 100M & 50.4 & 0.055 & 47.1 & 0.357 & 0.104 & 51.0 & 0.30 & 0.41 & -- \\
Reviser-Edit 100M & 57.6 & 0.125 & 46.1 & 0.266 & 0.283 & 50.6 & 0.05 & 0.31 & 0.592 \\
AR rewriter 300M & 54.8 & 0.075 & 49.7 & 0.398 & 0.187 & 53.4 & 0.26 & 0.48 & -- \\
Reviser-Edit 300M & 59.3 & 0.141 & 46.2 & 0.323 & 0.306 & 52.7 & 0.04 & 0.33 & 0.514 \\
\bottomrule
\end{tabular}}
\caption{Editing (greedy over action types; training seed 1, seed 2 in the text). SARI with one reference; EM: exact match. ERRANT F$_{0.5}$ on CoEdIT grammar items (485) and IteraTeR fluency items (88), our setup (not BEA-2019). JFLEG: GLEU on the test set with the four official references. Reference as input kept: on items whose reference differs from the input, we give the reference as the input with the same instruction and report the share returned unchanged (not measured for CoEdIT-large). Codex: preference for Reviser-Edit over the AR rewriter of the same size at matched instructions (300 items, both orders, tie $=0.5$; 95\% CI $\pm$0.05). Paired SARI differences (Edit $-$ AR, 95\% CI): CoEdIT $+7.2\,[+6.0,+8.2]$ (100M), $+4.5\,[+3.5,+5.5]$ (300M); IteraTeR $-1.0\,[-3.7,+1.6]$, $-3.4\,[-6.0,-0.8]$.}
\label{tab:edit}
\end{table*}

\paragraph{Edit vs.\ rewrite.} With the same data, steps and size (but different starting models), the edit layout beats the AR rewriter on CoEdIT SARI by 7.2 (100M) and 4.5 points (300M) and about twice its exact-match rate (\Cref{tab:edit}). Copying the input scores 57.9 on CoEdIT SARI; the lead comes from neutralization items, where the rewriter changes text it should keep (SARI 49.0/57.2 vs.\ 66.5/68.7; copying 69.1), while on grammar items both beat copying by 4--8 points. It is also better on fluency edits (F$_{0.5}$ 0.28--0.31 vs.\ 0.10--0.19) and on synthetic repairs (74.7/77.6 vs.\ 64.0/70.9 SARI). Codex prefers Reviser-Edit at 100M (0.592; sign test $p=0.0006$) and rates the two even at 300M (0.514). The AR rewriter is better at grammatical error correction (F$_{0.5}$ 0.36--0.40 vs.\ 0.27--0.32; Codex also prefers it on CoEdIT grammar items) and on IteraTeR at 300M (level at 100M), and it more often leaves the reference unchanged when given it as input. On JFLEG the two layouts tie (GLEU 50.6 vs.\ 51.0 at 100M, 52.7 vs.\ 53.4 at 300M; both paired CIs include 0), well above copying the input (40.5). A second training seed of both pairs gives the same results (CoEdIT SARI gap +7.4 at 100M, +3.6 at 300M; no SARI score moves by more than 1.3); JFLEG and Codex use seed 1. The synthetic-repair lead comes from repairs in the training mix (\Cref{app:editdata}).

\paragraph{Against public editors.} Reviser-Edit 300M beats the three zero-shot instruction-tuned LMs, which rewrite almost every input, on CoEdIT SARI (by 13--21 points) and on both ERRANT sets; only SmolLM2-360M is level with it on IteraTeR SARI and JFLEG (53.7 vs.\ 52.7). CoEdIT-large leads on all three sets (JFLEG 59.4 vs.\ 52.7), and Codex prefers it over Reviser-Edit 300M (0.260 [0.200, 0.322]); it is larger, instruction-tuned from Flan-T5 and trained on the CoEdIT data alone. No model here beats copying the input on IteraTeR SARI.

\paragraph{Cost.} Reviser-Edit writes only the changed tokens. On CoEdIT validation it needs a third of the AR rewriter's decoding steps (0.33$\times$), and on the GPU of \Cref{sec:results_speed} it has 0.35$\times$ the batch-1 latency and 2.3--2.7$\times$ the batch-32 throughput, with equal or lower peak memory. Analytic FLOPs fall less (0.63--0.66$\times$) because both models read the same instruction and draft. Part of the saving is doing less: Reviser-Edit leaves most paraphrase inputs nearly unchanged.

\section{Measured Cost of Writing}
\label{sec:results_speed}
On one RTX 2070 SUPER with the same KV-cached decoder for both models, a Reviser step costs the same as an AR step (26.6--27.1\,ms at 100M, 28.4--29.7\,ms at 300M, batch 1). Reviser needs 1.2--1.5 actions per kept token, so per kept token it is 1.56--1.67$\times$ (released), 1.32--1.35$\times$ (continued, 100M) and 1.21--1.26$\times$ (\Delete+RL) slower than AR at batch 1, and 1.5--3.0$\times$ at batch 32, where static batches wait for the longest trajectory. It finishes a prompt sooner (0.55--0.78$\times$ at batch 1) because it writes fewer tokens. Counted FLOPs agree: 1.31$\times$ AR with the prompt pass, 1.47$\times$ without it (\Cref{fig:efficiency}).

\section{Discussion}
\label{sec:discussion}
\paragraph{Where does the cursor pay off?} The cursor earns its cost where the right action is local and well defined: deleting a just-typed repeated token, and editing a draft, where unchanged text is copied for free. In continuation, the synthetic orders never tell the model \emph{why} to move; learning that from a reward is the natural next step. Even so, Reviser already writes at near-AR cost and beats diffusion LMs of its size, and one of three judges slightly prefers it to its AR twin at 100M.

\section{Conclusion}
\label{sec:conclusion}
Reviser writes and edits text with one cursor action per step at the cost of an AR step. Cursor edits beat a matched AR rewriter on CoEdIT, with a third of its steps and latency, and on fluency edits, though the AR rewriter makes more precise grammar corrections. In continuation, Reviser beats diffusion LMs of its size at near-AR cost and learns to fill gaps and delete repeated tokens. Real edit histories, learning when to move from a reward, and larger models are the next steps.

\section*{Limitations}
\textbf{Scale.} We train only 100M and 300M models and cannot say how the trade-offs change at larger scale. \textbf{Synthetic trajectories.} All training orders are synthetic; human-fitted orders are only fitted to aggregate statistics of CoAuthor sessions, and edit trajectories are diff-derived, not recorded keystrokes. \textbf{Budgets and seeds.} The released models and AR baselines come from the earlier version of this work and their budgets are not matched; ablation arms are single runs, and a rerun shows run-to-run differences of about 2 PPL. Continuation models have one training seed (RL: three seeds at 100M); the editing models have two. \textbf{Metrics.} The RL reward is a likelihood under a Qwen2.5 model and overlaps our perplexity metrics. LLM judges, even length-controlled and in both orders, are not humans, and we ran no human evaluation. ERRANT is computed with our own single-reference setup, not the BEA-2019 protocol. \textbf{History-only input.} The model reads its action history, not the current text, so after many moves it must infer the canvas from the history; we did not test giving it the canvas. \textbf{Operators.} \Delete{} is a single-token backspace. The continuation models use it only on the token just typed and mainly for the kind of mistake seen in training; Reviser-Edit uses it to remove draft text after moving to it. Replace and span operators are not trained.

\section*{Ethical Considerations}
The Reviser models are small research prototypes trained on English web text (FineWeb, C4, DCLM) and public edit corpora; they are not instruction-tuned for safety, and their outputs can reproduce inaccuracies and biases of web text. Part of the edit data was written by Qwen2.5-32B-Instruct and Claude Haiku 5.5; we release it only as permitted by the respective terms. LLM judges were run through the providers' standard interfaces on public prompts and model outputs. An editing model can make small, targeted changes that alter the meaning or attribution of a text, or paraphrase text to obscure its origin; we encourage work on edit-based generation to treat edit provenance as a first-class concern. We used AI coding assistants (Claude Code and Codex) to write and run experiment code and to edit the text of this paper; the author takes full responsibility for all code, results and text.

\bibliography{references}

\appendix

\section{Algorithms and Analytic Compute}
\label[appendix]{app:moved}
This appendix collects material that was in the main text of the earlier version.

\subsection{Compute Tradeoffs in NAR and Edit-Based Transformers}
\label[subappendix]{sec:why_nar_costs}
Many NAR and edit-based Transformers obtain flexibility or parallelism by repeating expensive sequence-level computation multiple times per response \citep{lee2018deterministic,ghazvininejad2019maskpredict,gu2019levenshtein,li2022diffusionlm,lou2024sedd,sahoo2024mdlm}. Under a shared FLOPs convention this often yields far higher total inference compute than a standard AR baseline. We report the multiplier
\[
I_{\text{infer}}^{\mathcal{M}} \;\triangleq\; \frac{F_{\text{infer}}^{\mathcal{M}}}{F_{\text{infer}}^{\text{AR}}},
\]
where $F_{\text{infer}}^{\mathcal{M}}$ is the total FLOPs to generate a length-$n$ response under method $\mathcal{M}$ (full derivations in \Cref{app:prior_art_efficiency_flops}). An AR baseline produces a length-$n$ response with one (KV-cached) full-sequence evaluation plus a vocab head per token. Most NAR families instead run $R$ full-sequence evaluations (refinement iterations, insertion rounds, or denoising steps), multiplying both the dominant trunk cost and the large vocab-head cost ($|V|\approx 50$k) by roughly $R$; e.g., CMLM commonly uses $R\approx 10$ and diffusion-style decoders use $R=T_{\text{dec}}$ from 10 to 2048.

The most compute-efficient NAR variants predict many (or all) tokens at once, but this creates a coordination problem: each position must agree with the unknown choices at other positions. One-shot NAR behaves like a product-of-marginals approximation $p(y\mid x)\approx\prod_{i=1}^{n} p(y_i\mid x)$ \citep{gu2018nonautoregressive}, which cannot reliably enforce inter-token constraints, manifesting as agreement errors, repetition, and contradictions, and motivating additional passes that raise compute.

Reviser avoids both issues: it predicts one cursor-relative action at a time conditioned on the full prior edit history (keeping global coherence through AR dependence while allowing non-monotonic edits), and each step is a standard next-action pass rather than a repeated full-sequence re-evaluation. If a fraction $p_{\text{move}}$ of actions are moves and the rest are inserts, producing $n$ tokens takes an expected $n_{\text{eff}}=n/(1-p_{\text{move}})$ steps, so
\[
I_{\text{infer}}^{\text{Reviser}}(n)\approx\frac{F_{\text{infer}}^{\text{AR}}(n_{\text{eff}})}{F_{\text{infer}}^{\text{AR}}(n)}.
\]
For $p_{\text{move}}\in\{0.20,0.28,0.33\}$ at $n=128$, $n_{\text{eff}}\in\{160,177.8,191.0\}$ and $I_{\text{infer}}^{\text{Reviser}}\in\{\num{1.25},\num{1.39},\num{1.50}\}$. \Cref{fig:efficiency} summarizes the analytic multipliers next to measured ones (\Cref{app:measured_flops}): at their default settings, refinement models need about 4--36$\times$ the compute of AR and diffusion LMs about 120--150$\times$. The earlier version used some settings that the released decoders do not use (LLaDA with 256 steps for 128 tokens, Dream only at 10--20 steps, one length candidate for Mask-Predict, early-exit heads for LevT, no caching for MDLM, and an encoder counted for NAT and Mask-Predict but not for their AR baseline); \Cref{fig:efficiency} and \Cref{app:prior_art_efficiency_flops} use the released settings. One-shot NAT and the balanced-tree Insertion Transformer are cheap but highly parallel, and such variants underperform because of the coordination problem above.

\subsection{Measured FLOPs}
\label[subappendix]{app:measured_flops}
To check the analytic estimates, we count the FLOPs that trained checkpoints actually execute. A PyTorch dispatch mode sees every executed operator with its real shapes and charges the dominant matrix multiplications (weight matmuls, the output head at every executed position, and attention); a multiply-add counts as 2 FLOPs, which does not change ratios. Every measurement is at batch 1 with the model's released decoder and is divided by an AR baseline that produces the same number of tokens on the same inputs, counting all stages (prompt or encoder passes included).
\begin{itemize}[leftmargin=*,itemsep=0.1em,topsep=0.2em]
\item \textbf{Reviser} (C4 continuation, 200 prompts, the released 100M Reviser vs.\ the 100M AR model): the AR baseline is charged its prompt pass plus its first $L$ decoding steps, with $L$ the Reviser's output length on that prompt. Reviser costs 1.31$\times$ AR [95\% CI 1.30, 1.32] including the prompt pass and 1.47$\times$ [1.45, 1.50] for the new tokens alone, at a move share of 0.316; the analytic proxy $1/(1-p_{\text{move}})$ gives 1.46.
\item \textbf{LevT and Mask-Predict} (WMT14 En--De, first 200 sentences of newstest2014, released fairseq checkpoints on CPU): the baseline is a same-shape encoder--decoder AR forced to emit each output. LevT with its default adaptive stopping (3.3 iterations on average) costs 4.18--4.20$\times$; Mask-Predict with 10 iterations costs 7.05$\times$ with one length candidate and 36.2$\times$ with the authors' default of five.
\item \textbf{SEDD and MDLM} (released OpenWebText checkpoints of one codebase with a shared 12-layer backbone, unconditional, 128 tokens, against the AR checkpoint of the same codebase): SEDD costs 33.6$\times$ ($T{=}32$) to 2086$\times$ ($T{=}2048$) per sampling attempt; its released sampler re-draws samples whose unigram entropy is low, which at 128 tokens raises the cost per accepted sample to 45.1--4432$\times$. MDLM with $T{=}1000$ costs 121$\times$ with caching and 1011$\times$ without; its default first-hitting sampler costs 129$\times$.
\item \textbf{Dream-7B, LLaDA-8B and SDLM} (GSM8K): with few-shot prompts (617 tokens on average) and 128 new tokens, against Qwen2.5-7B, Dream costs 10.8$\times$ ($T{=}10$), 21.6$\times$ ($T{=}20$) and 138$\times$ ($T{=}128$, its own setting for 128 tokens), and LLaDA 75.0$\times$ ($T{=}64$) and 150$\times$ ($T{=}128$; Qwen2.5-7B substitutes for the gated LLaMA3-8B baseline). With zero-shot chat prompts, SDLM-D4 at 3B and 32B costs 1.85--2.26$\times$ the matching Qwen2.5 AR model generating the same number of tokens with its confidence-threshold decoder (227--232$\times$ with its self-speculative mode).
\end{itemize}
The default settings (the 4--150$\times$ in \Cref{sec:intro}) are LevT with adaptive stopping, Mask-Predict with five length candidates, Dream and LLaDA with $T{=}128$, SEDD with $T{=}128$ (131$\times$ per attempt) and MDLM with first hitting. At the same settings, measured and analytic values agree for MDLM (121$\times$ vs.\ 122$\times$ with caching, 1011$\times$ vs.\ 1009$\times$ without) and SEDD (per attempt, within 5\%). Mask-Predict, Dream and LLaDA measure 12--25\% above the analytic values, which use the reference sizes and leave out the prompt (617 tokens for Dream and LLaDA), and LevT measures 4.2$\times$ (5.5$\times$ against a decoder-only AR, the convention of its analytic estimate) vs.\ 6.3$\times$. FLOPs are not wall-clock time: at batch 1 on a GPU, AR decoding is limited by memory bandwidth, so diffusion steps that process many positions per pass are relatively cheap in time. Dream and LLaDA at $T{=}128$ take 7.2$\times$ and 8.4$\times$ the AR time on GSM8K (one RTX 3090), and MDLM with first hitting 2.2$\times$; the parallel models are fastest at few steps, where their quality drops (Dream at $T{=}10$: 3\% GSM8K accuracy vs.\ 44\% for Qwen2.5-7B on the first 32 test questions).

\subsection{Decoding and Obfuscation--Restoration Algorithms}

\begin{algorithm}[t]
\caption{Reviser decoding.}
\label{alg:reviser_decode}
\small
\begin{algorithmic}[1]
\Require prefix $x=(x_1,\dots,x_m)$ (possibly empty); max steps $T_{\max}$; max canvas $L_{\max}$
\State \textbf{Phase 1 (prefix seeding):} $C\!\leftarrow\!()$, $u\!\leftarrow\!0$, $H\!\leftarrow\!()$
\For{$i=1,\dots,m$}
  \State $\Insert(x_i)$: $C\!\leftarrow\!(C,x_i)$, $u\!\leftarrow\!u{+}1$, $H\!\leftarrow\!(H,\Insert(x_i))$
\EndFor
\State $C_1\!\leftarrow\!C,\ u_1\!\leftarrow\!m,\ H_1\!\leftarrow\!H$
\State \textbf{Phase 2 (continuation decoding):}
\For{$t=1,2,\dots,T_{\max}$}
  \State $r_t\leftarrow f_\theta(H_t)$;\ \ $m_t\leftarrow\textsc{Mask}(C_t,u_t,L_{\max},|x|)$
  \State $a_t\sim\mathrm{softmax}(r_t+m_t)$
  \If{$a_t=\Stop$} \Return $C_t$
  \Else
    \State $(C_{t+1},u_{t+1})\leftarrow\textsc{Exec}(C_t,u_t,a_t)$
    \State $H_{t+1}\leftarrow(H_t,a_t)$
  \EndIf
\EndFor
\State \Return $C_{T_{\max}+1}$ \Comment{fallback if no \Stop}
\end{algorithmic}
\end{algorithm}

\begin{table*}[t]
\centering
\small
\setlength{\tabcolsep}{6pt}
\begin{tabular}{p{0.47\linewidth}p{0.47\linewidth}}
\toprule
\textbf{Obfuscation (target $\rightarrow$ blank)} & \textbf{Restoration (blank $\rightarrow$ target)} \\
\midrule
\begin{minipage}[t]{\linewidth}\vspace{0pt}
\begin{enumerate}[leftmargin=*,itemsep=0.2em]
\item Initial\ \texttt{[Hi,|how are you?]}
\item $\Delete$\ \texttt{[Hi|how are you?]}
\item $\Delete$\ \texttt{[|how are you?]}
\item $\Move(+4)$\ \texttt{[how are you?|]}
\item $\Delete$\ \texttt{[how are you|]}
\item $\Delete$\ \texttt{[how are|]}
\item $\Delete$\ \texttt{[how|]}
\item $\Delete$\ \texttt{[|]}
\end{enumerate}
\end{minipage}
&
\begin{minipage}[t]{\linewidth}\vspace{0pt}
\begin{enumerate}[leftmargin=*,itemsep=0.2em]
\item Initial\ \texttt{[|]}
\item $\Insert$(\texttt{how})\ \texttt{[how|]}
\item $\Insert$(\texttt{are})\ \texttt{[how are|]}
\item $\Insert$(\texttt{you})\ \texttt{[how are you|]}
\item $\Insert$(\texttt{?})\ \texttt{[how are you?|]}
\item $\Move(-4)$\ \texttt{[|how are you?]}
\item $\Insert$(\texttt{Hi})\ \texttt{[Hi|how are you?]}
\item $\Insert$(\texttt{,})\ \texttt{[Hi,|how are you?]}
\item $\Stop$\ \texttt{[Hi,|how are you?]}
\end{enumerate}
\end{minipage}
\\
\bottomrule
\end{tabular}
\caption{Side-by-side obfuscation and restoration trajectory for a full continuation without a prefix. Restoration reverses the obfuscation and inverts each step ($\Delete\leftrightarrow\Insert$ of the deleted token, $\Move(\Delta)\leftrightarrow\Move(-\Delta)$).}
\label{tab:trajectory}
\end{table*}

\paragraph{Obfuscation--restoration supervision.}
We construct training trajectories by corrupting a target and inverting the corruption (\Cref{alg:obf_restore}; a worked side-by-side example is in \Cref{tab:trajectory}). Initialize the editable canvas to the continuation target and the cursor at a uniformly random boundary: $C\leftarrow y$, $u\sim\mathrm{Uniform}(\{0,\dots,|y|\})$. At each obfuscation step we sample, with validity masking, $\Delete$ (probability $0.8$), which deletes the token immediately left of the cursor, or $\Move(\Delta)$ (probability $0.2$) with $\Delta$ uniform over valid moves; deleted tokens are recorded in temporal order. Moves use powers-of-two displacements $\Mset=\{\pm 2^k : 2^k\le \text{max\_move}\}$. Obfuscation terminates when the canvas is empty; trajectories that fail to empty within $K_{\max}$ steps are discarded and resampled.

Given the obfuscation sequence $B$ and the deleted-token stack $\mathcal{D}$, the restoration trajectory $\phA$ is built by scanning $B$ last-to-first and inverting each action: each $\Delete$ becomes $\Insert(x)$ with $x=\mathrm{pop}(\mathcal{D})$ (LIFO order), and each $\Move(\Delta)$ becomes $\Move(-\Delta)$; we then append $\Stop$. This strict inverse yields one restoration action per obfuscation action ($T_{\text{rest}}=n_{\text{obf}}+1$), and executing $\phA$ from a blank canvas deterministically reconstructs the target. Because the construction reverses executor dynamics, every training target is a valid next edit for the current canvas/cursor state rather than a label detached from state, which yields dense, state-consistent edit-history supervision using the same interface as inference.

\begin{algorithm}[t]
\caption{Obfuscation--restoration trajectory.}
\label{alg:obf_restore}
\small
\begin{algorithmic}[1]
\Require target $y=(y_1,\dots,y_n)$; move set $\Mset$; max steps $K_{\max}$
\State $C\leftarrow y$;\ $u\sim\textsc{Uniform}(\{0,\dots,n\})$;\ $B\leftarrow()$;\ $\mathcal{D}\leftarrow()$
\For{$k=1,\dots,K_{\max}$}
  \If{$|C|=0$} \textbf{break} \EndIf
  \State sample $b_k\sim\pi_{\text{obf}}$ ($\Delete$ w.p.\ $0.8$, else $\Move(\Delta)$), masked
  \If{$b_k=\Delete$}
    \State $x\leftarrow c_u$; delete $x$; $u\leftarrow u{-}1$; push $x$ to $\mathcal{D}$
  \Else\ $u\leftarrow u+\Delta$
  \EndIf
  \State append $b_k$ to $B$
\EndFor
\State $\phA\leftarrow()$
\For{$k=|B|,|B|-1,\dots,1$}
  \State if $b_k=\Delete$: append $\Insert(\mathrm{pop}(\mathcal{D}))$; else append $\Move(-\Delta)$
\EndFor
\State append $\Stop$ to $\phA$; \Return start state and restoration $\phA$
\end{algorithmic}
\end{algorithm}

\section{Judge Protocol}
\label[appendix]{app:judges}
\paragraph{Skywork.} Skywork-Critic-Llama-3.1-8B judges all 3,000 continuation pairs with its released pairwise prompt. \emph{Original protocol}: full texts, one order per example drawn with a fixed seed, no ties, unparseable verdicts dropped. \emph{Matched length}: both continuations are cut to $L=\min(n_X,n_Y)$ new GPT-2 tokens and judged in both orders; a split verdict counts as a tie (0.5). CIs: 95\% bootstrap over prompts. Skywork is strongly position-biased: its two orders agree on only 76--84\% of prompts against the public LMs (Codex: 93--99\%), which is why we always average both orders.

\paragraph{Codex and Claude (continuation).} Codex (CLI 0.156.0, model gpt-6-astra, reasoning effort high, read-only sandbox, subscription login, no tools) and Claude (Claude Code CLI, model claude-opus-5-5, all tools disabled) receive batches of items. Each item shows the 35-token prefix and two texts A and B, both cut to the same number of new tokens; the two orders go to separate calls, and system names are never shown. Items: 300 per pair for the models in \Cref{tab:main} (item seed 20261008) and for the public-LM pairs of \Cref{tab:lmcompare}; 100 per pair (50 C4, 50 DCLM; seed 20260924) for the ablations in \Cref{tab:ablate}. Score per item: win 1, tie 0.5, loss 0, averaged over the two orders; we report the mean with a 95\% bootstrap CI over items (10,000 resamples), an exact sign test, and position consistency. The instruction (verbatim):
\begin{quote}\small\ttfamily
You are judging text continuations. For each item you get a PREFIX (the start of a document) and two complete texts, A and B, each of which begins with (approximately) that prefix and continues it. Decide which text is the better continuation of the prefix, judging: coherence and consistency with the prefix, fluency and grammaticality, and absence of degenerate repetition or nonsense. Both texts may be cut off mid-sentence; do not penalize an abrupt ending, and do not prefer a text for being longer or shorter. Answer "tie" only if you truly cannot prefer one.
\end{quote}
followed by an output-format instruction (one JSON verdict per item).

\paragraph{Codex (editing).} 300 items (150 CoEdIT validation, 150 IteraTeR human test; stratified by task, seed 20261010), the same for both sizes; Reviser-Edit 300M vs.\ CoEdIT-large uses a 150-item sub-sample. Items whose two outputs are identical are scored 0.5 without a call. Same judge settings, scoring and orders as above. Instruction (verbatim):
\begin{quote}\small\ttfamily
You are judging text edits. Each item has an INSTRUCTION, an INPUT text, and two edited versions of the input, A and B, produced by two different systems. Decide which version better carries out the instruction on the input while keeping the meaning of the input (except where the instruction itself asks for a change of content). Judge whether the instruction was followed, whether the meaning was kept, and whether the result is fluent and grammatical. Do not prefer a version for being longer or shorter, or for changing more or less, except as the instruction requires. A version may be identical to the input; judge it like any other. Answer "tie" if both are equally good or equally bad.
\end{quote}
Against the AR rewriter, Codex's preference for Reviser-Edit is 0.592 [0.542, 0.640] at 100M and 0.514 [0.465, 0.562] at 300M (95\% intervals). Position consistency was 0.95--0.97 for all edit pairs; by task, Reviser-Edit loses on CoEdIT grammar items at both sizes (0.44, 0.37) and wins on neutralization (0.66, 0.67).

\section{Training Data and Trajectories}
\label[appendix]{app:editdata}
This appendix gives the recipes for the continuation trajectories of \Cref{sec:training,sec:results_ablate} and for the edit data.

\subsection{Continuation Trajectories}
\label[subappendix]{app:trajdata}

\paragraph{Texts and prompt.}
All continuation runs use the 2,640,539 FineWeb texts of the original run-and-jump run (\Cref{tab:main}). Each has 100--292 GPT-2 tokens: a 35-token prompt $x$ and a continuation $c=(c_1,\dots,c_m)$, $65\le m\le257$. A trajectory starts with 35 \Insert{} actions that write $x$, with masked labels. The generator works on $c$ alone, with the cursor counted from the end of the prompt, so no action enters or deletes the prompt. A trajectory ends with \Stop{} and has at most 350 actions in total.

\paragraph{Run-and-jump order (\Cref{alg:runjump}).}
A \emph{gap} is a maximal run of unwritten tokens, the unwritten tail included; its \emph{boundary} is the canvas position after the written tokens that precede it. In a \emph{run step}, with probability $\lambda=0.82$, the generator types the token that follows the one left of the cursor, if that token is unwritten. Otherwise it \emph{jumps}. It picks a gap with weight $(1+d)^{-1.5}$, where $d$ is the canvas distance from the cursor to the gap's boundary (the gap at the cursor has $d=0$). It moves there and types the gap's token at offset $s$; the $s$ tokens it skips become a hole that a later backward jump fills. For a gap of $L>1$ tokens the offset is drawn as follows (for $L=1$, $s=0$). In the unwritten tail, with probability $p_{\text{far}}=0.9$, $s$ is uniform on $\{0,\dots,L-1\}$ (a far start). Otherwise a skip happens with probability $p_{\text{skip}}=0.3$, and always in the gap at the cursor, where $s=0$ would only continue the run. A skip starts at $s=1$ and grows by one with probability $q=0.85$ while $s<L-1$; without a skip, $s=0$. A displacement $\Delta$ becomes moves $\pm2^k$ with the sign of $\Delta$, largest first ($-7\to-4,-2,-1$). Each token is written once. The dial $p_{\text{fill}}$ (ablations only) adds one rule: before a run step, if another gap exists, with probability $p_{\text{fill}}$ the generator jumps to one of the other gaps instead (same weights and offsets).

\paragraph{Orders, length and seeds.}
Our builder stores each order as a permutation $\pi$ of $\{1,\dots,m\}$ and turns it into actions: for each $j$ in turn, move to the position after the written tokens with index below $j$, then insert $c_j$. This reproduces the generator's actions, and the left-to-right, human-fitted and neural orders of \Cref{tab:ablate} are converted the same way. If a trajectory exceeds 350 actions, its order is redrawn, up to 200 times, and then written left to right (5 texts in the main data; 1.4\% at $\lambda=0.70$). The original generator, which made the reference data, dropped such texts instead; the 2.64M texts are those it used. Random streams are seeded per data shard (orders: 123; errors: 7).

\paragraph{Self-correction.}
Errors are drawn on top of a fixed order, with their own random stream. For each $c_j$, in writing order, with probability $\epsilon=0.05$ the writer types a wrong token $w$ in $c_j$'s slot, deletes it, and types $c_j$. The token $w$ is taken from a uniformly chosen continuation position $i$ with $0<|i-j|\le4$ whose token is neither $c_j$ nor end-of-text; if there is none, $w$ is a unigram sample (3,300 of 17.5M errors). The label of $w$ is masked; \Delete{} and the retyped $c_j$ are trained. If the errors push a trajectory past 350 actions, they are redrawn up to 200 times, and then the text's rate is halved until it fits (2.1\% of texts). Before training, the executor replays every trajectory and checks that it writes $x\,c$ exactly, ends with \Stop, keeps moves on the canvas, leaves the prompt unchanged, masks exactly the wrong inserts and has 100--350 actions. The main data (2.64M trajectories, mean 240 actions) have 0.330 moves (69\% backward) and 0.046 \Delete{} actions per continuation token; none of their 525M supervised labels is invalid under the prompt lock.

\begin{algorithm}[t]
\caption{Run-and-jump trajectory with self-correction (main recipe).}
\label{alg:runjump}
\small
\begin{algorithmic}[1]
\Require prompt $x_{1..35}$; continuation $c_{1..m}$; $\lambda{=}0.82$, $\alpha{=}1.5$, $p_{\text{skip}}{=}0.3$, $q{=}0.85$, $p_{\text{far}}{=}0.9$, $\epsilon{=}0.05$
\State $\phA\leftarrow(\Insert(x_1),\dots,\Insert(x_{35}))$ \Comment{labels masked}
\State $W\leftarrow\emptyset$;\ $u\leftarrow0$ \Comment{written set; cursor}
\While{$|W|<m$}
  \State $n\leftarrow W_{(u)}+1$ \Comment{$W_{(u)}$: $u$-th smallest; $W_{(0)}{=}0$}
  \If{$n\le m$, $n\notin W$, $\mathrm{rand}<\lambda$} \Comment{run step}
    \State \Call{Type}{$n$}
  \Else \Comment{jump}
    \State gaps $g$: maximal unwritten runs $c_{a_g..a_g+L_g-1}$
    \State boundaries $b_g\leftarrow|\{i\in W:i<a_g\}|$;\ $s\leftarrow0$
    \State sample $g$ with probability $\propto(1+|b_g-u|)^{-\alpha}$
    \If{$a_g{+}L_g{>}m$, $L_g{>}1$, $\mathrm{rand}<p_{\text{far}}$}
      \State $s\sim\textsc{Uniform}(\{0,\dots,L_g{-}1\})$ \Comment{far start}
    \ElsIf{$L_g{>}1$, ($b_g{=}u$ or $\mathrm{rand}<p_{\text{skip}}$)}
      \State $s\leftarrow1$ \Comment{skip}
      \State \textbf{while} $s<L_g{-}1$ and $\mathrm{rand}<q$: $s\leftarrow s{+}1$
    \EndIf
    \State append \Call{Pow2}{$b_g-u$};\ $u\leftarrow b_g$
    \State \Call{Type}{$a_g+s$} \Comment{$c_{a_g..a_g+s-1}$: a hole}
  \EndIf
\EndWhile
\State append \Stop; redraw if $|\phA|>350$; replay check
\Statex
\Procedure{Type}{$j$} \Comment{own random stream}
  \If{$\mathrm{rand}<\epsilon$}
    \State $w\leftarrow c_i$, $i$ uniform with $0<|i-j|\le4$,
    \Statex \hskip\algorithmicindent\hskip\algorithmicindent $c_i\ne c_j$, $c_i\ne$ end-of-text
    \State append $\Insert(w)$ (label masked), $\Delete$
  \EndIf
  \State append $\Insert(c_j)$;\ $W\leftarrow W\cup\{j\}$;\ $u\leftarrow u{+}1$
\EndProcedure
\Statex \textsc{Pow2}$(\Delta)$: moves $\pm2^k$ with the sign of $\Delta$, largest first ($-7\mapsto-4,-2,-1$).
\end{algorithmic}
\end{algorithm}

\paragraph{Ablation arms.}
Each arm continues the released 100M model for 83.9M tokens with the same texts, seed and learning-rate schedule; only the trajectories change. The rows of \Cref{tab:ablate}, in order:
\begin{itemize}[nosep,leftmargin=*]
\item \emph{Reference}: the run-and-jump model of \Cref{tab:main} (without \Delete), taken at the same token count; it was trained on the original generator's data.
\item \emph{Left to right only}: $\pi=(1,\dots,m)$.
\item \emph{$\lambda=0.98$, $0.95$, $0.90$, $0.86$} (four rows): only $\lambda$ changes.
\item \emph{$\lambda=0.82$, data rebuilt}: the reference recipe rebuilt by our builder (new draws, per-shard seeds, redraw instead of drop).
\item \emph{$\lambda=0.70$}: only $\lambda$ changes.
\item \emph{+ jumps between gaps}: the reference dials plus $p_{\text{fill}}=0.1$.
\item \emph{Human-fitted statistics}: run-and-jump with dials fitted to CoAuthor ($\lambda=0.993$, $\alpha=1.25$, $p_{\text{skip}}=q=0$, $p_{\text{far}}=0.05$, $p_{\text{fill}}=0.3$). A session's order sorts the GPT-2 tokens of its final text by the time they were completed (text deleted later is dropped; AI suggestions are kept). The dials were fitted by grid search to run lengths, jump rate and length, backward share and actions per token (held-out sessions, fitted vs.\ human: mean run 36.2 vs.\ 36.0 tokens, 2.14 vs.\ 2.16 jumps per 100 tokens).
\item \emph{Neural human order}: a pointer network, trained by teacher forcing on CoAuthor orders, picks the next token among the unwritten ones (a fine-tuned DistilRoBERTa reads the whole text; a two-layer decoder attends to past picks); sampled at temperature 1. On our texts it makes 4.6 jumps per 100 tokens (humans: 2.2).
\item \emph{Random order, 12\% junk inserts}: the reversed walk of \Cref{alg:obf_restore} on the continuation, starting at its end, with three operations: delete left of the cursor (0.70), one uniformly chosen valid move $\pm2^k$ (0.18), insert a uniformly random token (0.12). Junk inserts reverse to \Delete{} actions; inserting a junk token that is later deleted is masked. Texts over 350 actions use junk rate 0.06 (12\%) or 0 (1\%).
\item \emph{2\%, 5\% (main recipe), 10\%}: self-correction with $\epsilon=0.02$, $0.05$, $0.10$; the 5\% arm uses the data recipe of the 608M-token runs.
\item \emph{5\%, exact rerun}: the 5\% arm retrained from byte-identical data with the same seeds on another machine.
\item \emph{Fixed 1 / 3 tokens later}: $w$ stays while $k=1$ or $3$ more tokens are written; the writer then moves back, deletes $w$, types $c_j$ and goes on.
\item \emph{Fixed after the last token}: every $w$ stays until the text is complete; they are then fixed in the order they were made.
\item \emph{Unigram sample}: $w$ is drawn from the unigram distribution of continuation tokens.
\item \emph{GPT-2 sample}: $w$ is sampled from GPT-2 (top-50, temperature 1) given the true preceding text, with $c_j$ excluded.
\item \emph{Near, wrong insert not masked}: the 5\% arm with the label of $w$ trained.
\item \emph{5\% near errors, left to right}: the 5\% errors on the left-to-right order.
\end{itemize}
\Cref{tab:ablate} shows 20 of the 34 arms; its reference, human-fitted and neural rows are separate runs with the same setup. The 14 arms not shown are: no skips ($p_{\text{skip}}=0$), no far start ($p_{\text{far}}=0$), the 5\% arm with another error seed, fixes at the end of the current run, unmasked unigram errors, 5\% errors on $\lambda=0.86$, on $\lambda=0.90$ and on the human-fitted dials, three samplings of the neural order (calibrated keep-typing bias, temperature 0.7, temperature 1.3), and three per-text mixtures (half neural, half run-and-jump; a quarter or a half human-fitted, the rest run-and-jump).

\subsection{Edit Data}
\label[subappendix]{app:editmix}

\paragraph{Mix.} The build targets 15\% restoration, 15\% no change and 8\% synthetic repair as shares of draft-plus-target tokens; the five edit categories share the rest. As shares of the 4,006,336 episodes per pass, the mix is: restoration 14.8\% (C4 and DCLM, about half each), no change 17.2\% (C4 and DCLM sentences and paragraphs), synthetic repair 8.9\%, grammar 14.7\% (C4\_200M 70\%, LLM paragraphs 26\%, CoEdIT 4\%), fluency 10.9\% (LLM 61\%, real sentences 39\%), clarity 11.5\% (LLM 69\%, real 31\%), coherence 9.2\% (LLM 82\%, real 18\%) and polish 12.9\% (LLM 87\%, real sentences 12\%, IteraTeR documents 2\%). Real sentence pairs come from the CoEdIT and IteraTeR train splits and are assigned to a category by their edit intent. LLM pairs are edit drafts written from clean paragraphs by Qwen2.5-32B-Instruct-AWQ (Apache-2.0) and, for clarity, Claude Haiku 5.5 (\texttt{claude-haiku-5-5}; about 3.6\% of the training tokens); the training pair is draft $\to$ clean. Evaluation items were removed from all pools. Five input-only task tokens mark the category.

\paragraph{Episode types.}
\emph{Restoration}: the first 35\% of a C4 or DCLM text (rounded) is the prompt, with no task token, and the rest is written on a blank canvas by a reversed obfuscation walk. The walk starts with the cursor at the end and, until the canvas is empty, deletes the token left of the cursor (0.70), inserts a junk token (0.12) drawn from the token distribution of 2,000 random restoration texts, or moves (0.18) in a uniformly chosen valid direction by $d$ tokens with $P(d)\propto d^{-1.5}$, written as power-of-two moves; at the left edge a delete becomes a move. Junk inserts that are deleted later are masked. The AR layout writes the same tokens left to right.
\emph{No change}: a C4 or DCLM sentence or paragraph (most passed the clean-text check below) is both draft and target, its task token is drawn uniformly from the five, and the edit answer is a single \Stop{} (the AR rewriter copies the draft).
\emph{Synthetic repair}: the clean side of a CoEdIT or IteraTeR sentence pair gets 0, 1, 2 or 3 errors with probabilities 0.2, 0.5, 0.2 and 0.1. Each error replaces, deletes or inserts a token or swaps two adjacent tokens, with equal probability; new tokens come from the token distribution of 2,000 random texts of this pool. Errors are drawn per episode, so each pass sees new ones. The prompt is the grammar task token and ``Fix the errors in this text''.
\emph{Instructions}: real pairs keep their own. LLM drafts, C4\_200M pairs and no-change episodes get, with probability 0.5, an instruction drawn uniformly from the distinct real instructions of their category, and none otherwise.

\paragraph{LLM drafts.}
Source paragraphs (C4, DCLM) first pass a Qwen2.5-7B-Instruct clean-text check ($P(\text{clean})\ge0.9$) and a screen for odd word forms. The drafting prompt shows the LLM the clean paragraph, a description of the category, four real before/after examples, and some of the paragraph's sentences, each with one flaw type drawn from a weighted menu for the category (e.g.\ ``a wrong preposition''); the LLM returns each sentence's earlier draft with only that flaw. We splice the drafts back into the paragraph (paragraph pairs) or keep each changed sentence as its own pair (sentence pairs). Haiku's clarity drafts used the same prompts and went through the same filters and reviewer. Rule filters reject a draft whose number of changed spots is far from the number requested, whose largest spot exceeds a category limit (3--12 words), that keeps less than 80\% of the final's tokens, whose length ratio lies outside 0.85--1.2, or that changes a number, a negation or an antonym, among other checks. The reviewer, Qwen2.5-32B-Instruct-AWQ (also the main drafting model), sees draft and final as texts A and B with the changed spots listed and is asked which is better written there; we read P(final is better) from its first-token probabilities. The first batch (grammar, coherence and polish paragraphs) was scored in both orders and kept if the mean was at least 0.7. Later batches were scored only with the final as text A and kept if $P\ge0.4$, the lowest value at which the mean can still reach 0.7. In the later batches, coherence drafts also had to pass a yes/no check that the change is a coherence edit ($P(\text{yes})\ge0.5$). Of 3.34M drafting prompts, 72\% passed the rule filters and the reviewer kept 55\% of those, so 1.33M drafts (40\%) were kept before deduplication and removal of evaluation overlap. In the later batches, the reviewer rejected 13\% of the filter-passing grammar paragraphs and 86\% of the clarity sentences drafted by Qwen.

\paragraph{Episodes.} \texttt{[TASK] instruction SEP draft EDIT actions}; the loss covers only the actions. Edit layout: the canvas starts as the draft with the cursor at its end; hunks from \texttt{difflib} are edited right to left (move to the hunk end, \Delete{} its old tokens, insert the new ones), then \Stop. AR layout: the prompt additionally contains ``\texttt{:}'' and the draft, the canvas starts empty, and the target is written left to right. Both layouts share the action vocabulary and see the same examples in the same order (data seed 1001), 187,797 steps of 64 episodes, AdamW (0.9, 0.95; weight decay 0.1), learning rate $5\times10^{-5}$ with 300 warmup steps and cosine decay to 10\%, fp16 mixed precision, maximum length 512, on 2$\times$RTX 4090 per run (5.6--13.2\,h; \$41.70 for the four seed-1 runs). Labels: Reviser-Edit 247M (110.9M inserts, 42.1M deletes, 82.3M moves, 12.0M stops; 269M before 22M masked inserts are removed), AR 531M.

\paragraph{A smaller mix.} An earlier pair of runs per size used the same recipe for 53,851 steps on a smaller mix without the synthetic-repair pool. The runs reported here are within 1.6 SARI of it on CoEdIT and IteraTeR for both layouts and sizes (edit: $-0.7$/$-1.6$ at 100M, $-0.1$/$-1.4$ at 300M; AR: $-0.4$/$-1.1$, $+0.5$/$+0.7$) and 8--23 SARI higher on synthetic repairs. With the smaller mix the edit$-$AR gap was about the same on CoEdIT ($+7.5$ at 100M, $+5.1$ at 300M) but negative on synthetic repairs ($-2.9$, $-3.6$). Longer training on more data thus raised only the in-domain synthetic scores.

\section{Reinforcement Learning Fine-Tuning}
\label[appendix]{sec:rl}
Generation is an episodic MDP: the state is $(C_t,u_t,H_t)$ with its validity mask (including the prompt lock), the actions are the edit operators, and transitions are given by the deterministic executor. The \emph{+RL} models in \Cref{tab:main} are fine-tuned from the self-correction model with a group-relative policy gradient \citep{shao2024deepseekmath}. Each step samples 16 prompts (35-token prefixes of FineWeb texts from the continuation training data, not from the C4 or DCLM evaluation sets) and 8 continuations per prompt with the evaluation sampler (mask, temperature 0.9, top-$k$ 50, at most 220 new actions plus \Stop). The terminal reward is
\begin{align*}
r={}&-\mathrm{NLL}_{\text{Qwen}}(y\mid x)-8\,\mathrm{rep}_4(y)\\
&-\max\!\Big(0,1-\tfrac{n_{\text{new}}}{150}\Big)-2\max\!\big(0,f-\mu(y)\big),
\end{align*}
where $\mathrm{NLL}_{\text{Qwen}}$ is the mean per-token NLL of the continuation given the prompt under Qwen2.5-3B-Instruct (plain text, no chat template; an empty continuation gets $-10$), $\mathrm{rep}_4=1-\#\text{distinct}/\#\text{total}$ 4-grams, $n_{\text{new}}$ is the number of new tokens, $\mu$ the moves per new token, and $f$ the batch mean of $\mu$ at the first step (0.143 at 100M seed 3, 0.149 at 300M). Advantages are $(r-\bar r_g)/(\sigma_g+10^{-4})$ within each group of 8. The loss is $-A\log\pi_\theta(a\mid s)$ summed over the sampled actions plus $0.2\cdot\mathrm{KL}(\pi_\theta\,\|\,\pi_{\text{ref}})$ computed exactly over the allowed top-$k$ set against the frozen starting model, divided by the number of generated actions; one on-policy update per batch, no ratio clipping, no value head. Optimization: 200 steps, learning rate $5\times10^{-6}$ (10 warmup steps), gradient clipping 1.0, bf16, one RTX 4090 ($\sim$19\,min at 100M, 29\,min at 300M). We ran three seeds at 100M and one at 300M.

Because the reward is a language-model likelihood, the GPT-2 and especially the Dream-7B (Qwen2.5-based) perplexities of the +RL models partly measure the training reward, and $\mathrm{rep}_4$, length, stopping and moves per token are reward terms. The LLM judges are independent of the reward, and MAUVE only partly (its features are BERT likelihoods and length); Skywork's matched-length score did not change with RL (53.0 before, 52.1--54.4 after).

\section{Full-Length Results of the Earlier Protocol}
\label[appendix]{app:fulllength}
For reference, \Cref{tab:quality} reproduces the full-length quality results of the earlier version of this work, computed on stored outputs of a decoder without the prompt lock; its full-length win rates against the diffusion and public LMs are the ``Full'' column of \Cref{tab:lmcompare}. Texts were not length-matched, Reviser texts were 66--94 tokens long and AR texts 216--220, and each pair was judged in one random order. \Cref{tab:main} replaces these numbers: it re-decodes every model with the prompt lock, so its texts differ. The full-length Skywork win rate of Reviser against AR is 61.3\%/54.4\% (100M/300M) as reported in the earlier version, 61.5\%/54.5\% when we re-judge the same stored texts, and 62.5\%/53.2\% on the new decodes of \Cref{tab:main}; on full texts the continued models score 54.0 (run-and-jump), 53.6 (+\Delete) and 53.8 (+RL, three seeds) at 100M, and 47.9 (+\Delete) and 48.4 (+RL) at 300M. Likewise, the AR rows of \Cref{tab:quality} score the earlier stored AR texts (GPT-2 evalPPL 13.52/12.51; the new decodes give 13.40/12.63), and its MAUVE values come from the earlier MAUVE harness, which is not comparable with the one used in \Cref{tab:main}.

\paragraph{Correction: repeated prompt in stored outputs.}
The decoder behind the results below did not enforce the prefix lock, so Reviser sometimes edited the prompt, and storing the prompt followed by the final canvas then repeated the prompt in 405 of 3000 Reviser 100M outputs (13.5\%) and 127 of 3000 Reviser 300M outputs (4.2\%; AR outputs are unaffected). The repeat lowered Reviser's GPT-2 evalPPL (100M: 30.64 vs.\ 33.52 without it; 300M: 17.77 vs.\ 18.15) and raised its MAUVE (re-implemented harness, not comparable to \Cref{tab:quality}: 0.863 vs.\ 0.782 at 100M, 0.901 vs.\ 0.895 at 300M; AR 0.890/0.830); arena win rates barely move (61.5\%/54.5\% with it, 61.8\%/54.8\% without). Dream-7B is unaffected by the repeat but, on prompt-edited examples, scores the model's easy-to-predict copy of the prompt; prefix-locked re-decodes score 25.68 at 100M (22.96 unlocked) and 16.03 at 300M (8 MC samples). \Cref{tab:quality} uses the corrected values.

\begin{table}[t]
\centering
\small
\setlength{\tabcolsep}{2.5pt}
\maxwidthbox{%
\begin{tabular}{lcccc}
\toprule
 & \multicolumn{3}{c}{evalPPL $\downarrow$} & \\
\cmidrule(lr){2-4}
Model & GPT-2 & GPT-2 L & Dream-7B & MAUVE $\uparrow$ \\
\midrule
AR Baseline 100M & \num{13.5210} & \num{13.79}$^\dagger$ & \num{13.9584} & \num{0.9038} \\
Reviser 100M & \num{33.52}$^\dagger$ & \num{34.68}$^\dagger$ & \num{25.68}$^\dagger$ & \num{0.8392}$^\ddagger$ \\
SEDD Small 169M & \num{158.7165} & -- & \num{117.0477} & \num{0.1798} \\
MDLM 170M & \num{144.6948} & -- & \num{94.3026} & \num{0.1509} \\
\midrule
AR Baseline 300M & \num{12.5122} & \num{11.40}$^\dagger$ & \num{12.0203} & \num{0.9103} \\
Reviser 300M & \num{18.15}$^\dagger$ & \num{18.21}$^\dagger$ & \num{16.03}$^\dagger$ & \num{0.9349}$^\ddagger$ \\
SEDD Absorb 353M & \num{98.6821} & -- & \num{83.2764} & \num{0.3594} \\
\bottomrule
\end{tabular}%
}
\caption{Earlier protocol, kept for reference; \Cref{tab:main} replaces these numbers (new decodes with the prompt lock and a different MAUVE harness, so its values differ, e.g.\ AR 100M MAUVE 0.935 there vs.\ 0.90 here). Quality on C4 continuation: evalPPL (lower better) under AR scorers (GPT-2 base and GPT-2 Large) and a diffusion scorer (Dream-7B), and MAUVE (higher better; shared 3k protocol). The GPT-2 column was previously mislabeled GPT-2 Large. $^\dagger$New measurement (our re-implemented harness reproduces the original GPT-2 values to within 0.005). Reviser GPT-2 values exclude the repeated prompt (previously 30.64/17.77); Reviser Dream-7B values are from prefix-locked re-decodes with 8 MC samples (previously 22.67/15.77; other rows: 128). $^\ddagger$Computed on stored outputs that repeat the prompt; not re-measured in this protocol (see text). --: not measured.}
\label{tab:quality}
\end{table}

\section{What the Self-Correction Models Delete}
\label[appendix]{app:deletes}
\paragraph{Planted mistakes.} We take 400 states from FineWeb texts of the run-and-jump training pool (a run-and-jump trajectory cut at a random point after the prompt; the ablation runs never saw these texts; the 608M-token runs read 96\% of the pool, each text at most once, so they have probably seen them, but the 84M-token run of the same recipe, which never saw them, gives 0.82 on repeated tokens (\Cref{tab:ablate}), close to their 0.85--0.87), insert one token at the cursor as if the model had just typed it, and compute the probability that the next action is \Delete{} under the decoding sampler (mask, temperature 0.9, top-$k$ 50). \Cref{tab:deletes} gives the result by kind of mistake. Sampling five more actions removes the planted token only when the first action does.

\begin{table}[t]
\centering
\footnotesize
\setlength{\tabcolsep}{2.5pt}
\maxwidthbox{%
\begin{tabular}{@{}l rrrr@{}}
\toprule
 & \multicolumn{2}{c}{100M} & \multicolumn{2}{c}{300M} \\
\cmidrule(lr){2-3}\cmidrule(lr){4-5}
Planted token & +\Delete & +RL (3 seeds) & +\Delete & +RL \\
\midrule
none, run-and-jump & 0.039 & 0.036--0.042 & 0.036 & 0.038 \\
none, left to right & 0.020 & 0.019--0.022 & 0.012 & 0.011 \\
repeat of left token & 0.845 & 0.856--0.861 & 0.870 & 0.864 \\
token from $\pm$4 (trained) & 0.515 & 0.514--0.532 & 0.532 & 0.520 \\
random token & 0.144 & 0.138--0.166 & 0.179 & 0.186 \\
GPT-2 small sample & 0.051 & 0.048--0.049 & 0.054 & 0.050 \\
\bottomrule
\end{tabular}}
\caption{Probability that the next action is \Delete{} after a planted token (400 states; 95\% CIs $\pm$0.01--0.04). The 100M model trained on the same order without self-correction gives 0 in every row. A token from within $\pm$4 positions is deleted with probability 0.80--0.82 when it also appears among the four tokens to its left and 0.30--0.32 otherwise. 46\% of the GPT-2 samples are likelier under GPT-2 than the reference token, so many are acceptable words; those with GPT-2 probability below 0.01 are deleted with 0.07--0.09.}
\label{tab:deletes}
\end{table}

\paragraph{Free writing.} In the 3,000 C4 continuations of \Cref{tab:main}, the self-correction models delete 0.026--0.033 times per new token (their training data has 0.046), and 93--97\% of outputs contain a delete. All but 15 of 68,526 deletes remove the token typed just before; no prompt token is deleted, and at least 99.9\% of deletes are followed at once by an insert in the same place. Only 7.2--8.5\% of deleted tokens repeat the token to their left (12.4\% in training); the most deleted tokens are frequent ones (``the'', ``.'', ``,'', ``and''). Under GPT-2 small, deleted tokens have median probability 0.002 (68\% below 0.01), against 0.085 for the tokens the models keep (23\% below 0.01), and the replacement is likelier than the deleted token in 68--71\% of deletes; 6\% of deletes retype the same token. Because deleted tokens are improbable to begin with, a fresh token is likely to score higher, so this is a local clean-up rather than evidence of better text. The final texts are not less repetitive: against the same continued training without self-correction (100M), adjacent repeated words tie and the share of repeated 4-grams rises by 0.6 points at matched length (95\% CI $[+0.15,+1.07]$), while RL lowers it by about 3 points at both sizes.

\section{Other Edit Operators}
\label[appendix]{sec:optional_ops}
The executor and the validity mask extend to richer operators, which we have not trained or tested and list as future work. \Replace$(x)$ overwrites the token left of the cursor (\Delete{} and \Insert$(x)$ in one action); \SpanDelete$(k)$ removes the $k$ tokens left of the cursor; \Swap{} exchanges the two tokens left of the cursor; \SpanMove$(k,j)$ moves the $k$ tokens left of the cursor to boundary $j$; and \SpanCopy$(k,j)$ copies them there. Each is masked to the states where it is valid (e.g.\ $1\le k\le u_t$) and has a natural inverse, so obfuscation--restoration trajectories can supervise it in the same way as \Insert, \Move{} and \Delete.

\section{Prior Work Mechanisms and FLOPs-Based Efficiency Accounting}
\label[appendix]{app:prior_art_efficiency_flops}

\subsection{Goal and definitions (FLOPs-based, shared constants across models)}
We quantify inefficiency by comparing inference FLOPs to produce a full length-$n$ output under each method's decoding procedure, relative to an autoregressive (AR) baseline.

We fix the same backbone hyperparameters for every method in Groups 1, 2 and 4 (so ratios are meaningful); Group 3 keeps each large model at its native size:
\[
n=128,\quad |V_c|=50{,}000,\quad d=d_{\text{model}}=768,\quad L=12,\quad d_{\text{ff}}=4d,\quad K_{\max}=16.
\]
For Reviser, the action vocabulary size is $|V_a| = |V_c| + |\Mset| + 2$ (the two extra actions are \Delete{} and \Stop); since $|\Mset|\ll |V_c|$, we often use $|V_a| \approx |V_c|$ when it does not materially affect ratios.

We count dominant matrix-multiply compute and omit small terms (biases, layernorm, activations, elementwise ops).
All FLOPs expressions in this appendix are built from primitives in \Cref{app:flops_primitives}.
We count one fused multiply-accumulate (MAC) as one unit in $F_{\text{mult}}(a,b,c)=abc$.
(To convert to conventions where one MAC equals two FLOPs, multiply all reported $F(\cdot)$ values by $2$; ratios are unchanged.)

For each method $\mathcal{M}$, define $F_{\text{infer}}^{\mathcal{M}}(n)$ as the total FLOPs required to output an entire length-$n$
sequence under that method's inference procedure (summing all model calls, scoring/rounding steps, termination checks, etc.).
We define inefficiency as $I_{\text{infer}}^{\mathcal{M}} \triangleq \frac{F_{\text{infer}}^{\mathcal{M}}(n)}{F_{\text{infer}}^{\text{AR}}(n)}$.

\subsection{Primitive FLOPs functions}
\label[subappendix]{app:flops_primitives}

Let $A\in\mathbb{R}^{a\times b}$ and $B\in\mathbb{R}^{b\times c}$. We define
$F_{\text{mult}}(a,b,c) \triangleq abc$, i.e., one multiply-accumulate (MAC) is one unit of compute.

We count dominant matrix-multiply compute and omit small terms (biases, layernorm, activations, elementwise ops). Here $L$ denotes Transformer depth (number of layers), while $n$ denotes sequence length. Because attention structure differs across settings, we distinguish full (bidirectional) attention from causal
(lower-triangular) attention.

Define $F_{\text{transformer}}^{\text{full}}(n)$ as the dominant FLOPs for one forward pass through an $L$-block Transformer
on a length-$n$ sequence with full self-attention:
\[
F_{\text{transformer}}^{\text{full}}(n)
\triangleq
L\Big(
4\,F_{\text{mult}}(n,d,d)
\;+\;
2\,F_{\text{mult}}(n,n,d)
\;+\;
2\,F_{\text{mult}}(n,d,d_{\text{ff}})
\Big).
\]
With $d_{\text{ff}}=4d:$
\[
F_{\text{transformer}}^{\text{full}}(n)=L\left(12nd^2 + 2n^2d\right).
\]

Causal self-attention uses only the $n(n+1)/2$ lower-triangular query--key pairs.
We therefore model the attention-score and attention-apply matmuls by replacing the full $n^2$ pair count with $n(n+1)/2$. Define the dominant FLOPs for one forward pass with causal self-attention as
\[
F_{\text{transformer}}^{\text{causal}}(n)
\triangleq
L\Big(
4\,F_{\text{mult}}(n,d,d)
\;+\;
n(n+1)\,d
\;+\;
2\,F_{\text{mult}}(n,d,d_{\text{ff}})
\Big),
\]
where $n(n+1)d$ comes from summing the triangular costs of $QK^\top$ and $\mathrm{Attn}\cdot V$, each equal to $\frac{n(n+1)}{2}d$. With $d_{\text{ff}}=4d:$
\[
F_{\text{transformer}}^{\text{causal}}(n)
=
L\left(12nd^2 + n^2d + nd\right).
\]

Under this triangular convention, a causal forward pass at length $n$ (computed in parallel under a causal mask) and KV-cached
autoregressive decoding up to token $n$ have the same dominant attention-matmul count.

Encoder self-attention is bidirectional, so we use the full-attention expression
$F_{\text{enc}}(n)\triangleq F_{\text{transformer}}^{\text{full}}(n)$.

We model one decoder cross-attention layer (dominant matmuls) between a target length-$n$ sequence and a source length-$n_{\text{src}}$
sequence as:
\begin{align*}
F_{\text{xattn,layer}}(n,n_{\text{src}})
&\triangleq
\underbrace{F_{\text{mult}}(n, d, d)}_{\text{Q proj (tgt)}}
+\underbrace{2F_{\text{mult}}(n_{\text{src}}, d, d)}_{\text{K,V proj (src)}}
+\underbrace{F_{\text{mult}}(n, d, d)}_{\text{O proj (tgt)}}\\
&\phantom{{}\triangleq{}}
+\underbrace{F_{\text{mult}}(n, d, n_{\text{src}})}_{\text{attn logits }(QK^\top)}
+\underbrace{F_{\text{mult}}(n, n_{\text{src}}, d)}_{\text{attn apply }(\mathrm{Attn}\cdot V)}.
\end{align*}
Stacking across $L$ decoder layers gives
$F_{\text{xattn}}(n,n_{\text{src}})\triangleq L\,F_{\text{xattn,layer}}(n,n_{\text{src}})$.

We model a dense vocab projection over $m$ positions as
$F_{\text{vocab}}(m)\triangleq F_{\text{mult}}(m,d,|V_c|)$.

\section{Model-by-model FLOPs accounting}

\subsection{Group 1. Anchors}
\subsubsection{Autoregressive Transformer (AR baseline)}
We model the total dominant matmul compute of KV-cached AR decoding up to length $n$ using the causal (triangular) attention convention.
Across $n$ decoding steps, the vocab head is applied once per generated token (total $n$ applications), yielding:
\[
F_{\text{infer}}^{\text{AR}}(n)
\;\triangleq\;
F_{\text{transformer}}^{\text{causal}}(n) + F_{\text{vocab}}(n).
\]

Using $n=128$, $|V_c|=50{,}000$, $d=768$, $L=12$, and $d_{\text{ff}}=4d$:
\[
F_{\text{transformer}}^{\text{causal}}(128)\approx \num{11.0238}\,\G,
\qquad
F_{\text{vocab}}(128)\approx \num{4.9152}\,\G,
\]
so
\[
F_{\text{infer}}^{\text{AR}}(128)\approx \num{15.939}\,\G.
\]

By definition, $I_{\text{infer}}^{\text{AR}}=1$.

\subsubsection{Cursor-style Edit Generation: Reviser (this work)}
\label[subsubappendix]{app:reviser_flops}

Reviser treats text generation as an editing process over a cursor position. At each step it predicts exactly one action (e.g., $\textsc{Insert}$(token), or $\textsc{Move}(\Delta))$, applies that edit to the current sequence, and continues iterating until termination. Unlike prior edit transformers that emit large per-position edit heads, Reviser emits a single action per step with a normal-sized token vocabulary head (plus a small set of edit actions), and it is trained with standard supervised learning (teacher forcing) in the same way as a typical Transformer.

Let $p_{\text{move}}$ be the fraction of actions that are cursor-moves. To produce $n$ inserted tokens, the expected number of action steps is
$n_{\text{eff}}=\frac{n}{1-p_{\text{move}}}$
(inserts and moves only; with $M$ moves and $D$ deletes, $n$ kept tokens take $n+M+2D$ steps plus \Stop).
For $n=128$, we report three settings: $p_{\text{move}}\in\{0.20,0.28,0.33\}$, giving
$n_{\text{eff}}\in\{160,\;177.78,\;191.04\}$.

We approximate Reviser's total decoding compute by evaluating the AR causal proxy at the effective length:
\[
F_{\text{infer}}^{\text{Reviser}}(n)\triangleq
F_{\text{transformer}}^{\text{causal}}(n_{\text{eff}})+F_{\text{vocab}}(n_{\text{eff}}).
\]

\[
I_{\text{infer}}^{\text{Reviser}}(n)
\triangleq
\frac{F_{\text{infer}}^{\text{Reviser}}(n)}{F_{\text{infer}}^{\text{AR}}(n)}.
\]

Using $F_{\text{infer}}^{\text{AR}}(128)\approx \num{15.939}\,\G$, we obtain:
\[
\begin{aligned}
p_{\text{move}}=0.20:\quad
&F_{\text{transformer}}^{\text{causal}}(160)\approx \num{13.8269}\,\G,\;
F_{\text{vocab}}(160)\approx \num{6.144}\,\G,\\
&F_{\text{infer}}^{\text{Reviser}}(128)\approx \num{19.9709}\,\G,\;
I_{\text{infer}}^{\text{Reviser}}\approx \num{1.253}.\\[0.25em]
p_{\text{move}}=0.28:\quad
&F_{\text{transformer}}^{\text{causal}}(177.78)\approx \num{15.3924}\,\G,\;
F_{\text{vocab}}(177.78)\approx \num{6.8267}\,\G,\\
&F_{\text{infer}}^{\text{Reviser}}(128)\approx \num{22.2191}\,\G,\;
I_{\text{infer}}^{\text{Reviser}}\approx \num{1.394}.\\[0.25em]
p_{\text{move}}=0.33:\quad
&F_{\text{transformer}}^{\text{causal}}(191.04)\approx \num{16.5644}\,\G,\;
F_{\text{vocab}}(191.04)\approx \num{7.3361}\,\G,\\
&F_{\text{infer}}^{\text{Reviser}}(128)\approx \num{23.9006}\,\G,\;
I_{\text{infer}}^{\text{Reviser}}\approx \num{1.4995}.
\end{aligned}
\]

\subsection{Group 2. Diffusion NAR LMs (100M)}

Many diffusion-style language models perform generation as $T_{\text{dec}}$ repeated full-sequence Transformer forward passes, each followed by a vocabulary projection. For methods with this structure, we define the generation FLOPs as
\[
F_\text{diffusion}^{\mathcal{M}_{\text{diff}}}(n,T_{\text{dec}})
\triangleq
T_{\text{dec}}\left(F_\text{transformer}^\text{full}(n) + F_\text{vocab}(n)\right).
\]

\subsubsection{SEDD}
\label[subsubappendix]{app:sedd_100m}

\citet{lou2024sedd} parameterize SEDD, a reverse discrete diffusion process, and generate by running a fixed number
$T_{\text{dec}}$ of ``network evaluations'' (function evaluations) of Transformer conditioned on time and noise.

A conservative matmul proxy is that each step computes full-sequence hidden states (full attention)
and produces categorical scores for all $n$ positions:

\[
F_{\text{infer}}^{\text{SEDD}}(n, T_{\text{dec}})
\triangleq F_\text{diffusion}^\text{SEDD}(n, T_{\text{dec}})
\]

We report two representative step counts:
\[
T_{\text{dec}}=32:\quad
F_{\text{infer}}^{\text{SEDD}}(128, 32)\approx \num{514.8424}\,\G,
\quad
I_{\text{infer}}^{\text{SEDD}}=\num{32.3008}.
\]
\[
T_{\text{dec}}=2048:\quad
F_{\text{infer}}^{\text{SEDD}}(128, 2048)\approx \num{32949.9154}\,\G,
\quad
I_{\text{infer}}^{\text{SEDD}}=\num{2067.2497}.
\]

\subsubsection{MDLM}
\label[subsubappendix]{app:mdlm_100m}

\citet{sahoo2024mdlm} introduce MDLM, a masked discrete diffusion language model. Sampling starts from an all-\texttt{[MASK]}
sequence and runs a discretized reverse diffusion with $T_{\text{dec}}$ steps; at each step, the model predicts token
distributions conditioned on the current partially denoised sequence.

Using the same conservative ``full logits per step'' proxy:
\[
F_{\text{infer}}^{\text{MDLM}}(n, T_{\text{dec}})
\triangleq F_\text{diffusion}^\text{MDLM}(n, T_{\text{dec}})
\]

\[
F_{\text{infer}}^{\text{MDLM}}(128, 1000)\approx \num{16088.8259}\,\G,
\quad
I_{\text{infer}}^{\text{MDLM}}=\num{1009.3993}.
\]
The released sampler caches: it re-runs the network only after some token has been unmasked, plus one final noise-removal pass. With unmasking times spread uniformly over the $T_{\text{dec}}$ steps, the expected number of network calls is $T_{\text{dec}}\big(1-(1-1/T_{\text{dec}})^{n}\big)+1\approx \num{121.2}$ at $n=128$, $T_{\text{dec}}=1000$, so
\[
F_{\text{infer,cache}}^{\text{MDLM}}(128, 1000)\approx \num{1950.0134}\,\G,
\quad
I_{\text{infer,cache}}^{\text{MDLM}}=\num{122.3422}.
\]
(The first-hitting sampler unmasks one token per call, $n+1$ calls: $I=\num{130.2125}$.)

\subsubsection{D3PM}
\label[subsubappendix]{app:d3pm_100m}

\citet{austin2021structured} introduce D3PM, a general discrete denoising diffusion probabilistic model. In the text setting, it corresponds to a
time-conditioned denoiser run for $T_{\text{dec}}$ discrete reverse steps to transform a highly corrupted sequence into a clean sample.

\[
F_{\text{infer}}^{\text{D3PM}}(n, T_{\text{dec}})
\triangleq
F_\text{diffusion}^\text{D3PM}(n, T_{\text{dec}})
\]

\[
F_{\text{infer}}^{\text{D3PM}}(128, 1000)\approx \num{16088.8259}\,\G,
\quad
I_{\text{infer}}^{\text{D3PM}}=\num{1009.3993}.
\]

\subsubsection{Diffusion-LM}
\label[subsubappendix]{app:diffusionlm_100m}

\citet{li2022diffusionlm} propose Diffusion-LM, which performs continuous diffusion over a length-$n$ sequence of vectors and runs a (bidirectional)
Transformer denoiser for $T_{\text{dec}}$ reverse steps. To convert continuous vectors to discrete tokens,
we upper-bound discretization/rounding by a dense vocab scoring matmul.

Let $|V|$ denote the (discrete) output vocabulary size for this model. We define
\[
F_{\text{round}}(n)\triangleq F_{\text{mult}}(n,d,|V|),
\]
(which equals $F_{\text{vocab}}(n)$ under our shared 100M configuration where $|V|=|V_c|$).

Each reverse step runs the denoiser once over the full length-$n$ sequence (full attention):
\[
F_{\text{denoise}}(n)\triangleq F_{\text{transformer}}^{\text{full}}(n).
\]

The clamping trick optionally applies rounding to the predicted $x_0$ on a subset of the
non-final reverse steps. Let $\gamma_{\text{clamp}}\in[0,1]$ denote the fraction of the first
$T_{\text{dec}}-1$ steps on which clamping (rounding) is applied.

We pay $T_{\text{dec}}$ denoiser evaluations, plus rounding once at the end, plus clamping-rounding on
$\gamma_{\text{clamp}}(T_{\text{dec}}-1)$ non-final steps:
\[
F_{\text{infer}}^{\text{DiffLM}}(n)
\triangleq
T_{\text{dec}}\,F_{\text{denoise}}(n)
+
\Big(1+\gamma_{\text{clamp}}(T_{\text{dec}}-1)\Big)\,F_{\text{round}}(n).
\]

Using $F_{\text{transformer}}^{\text{full}}(128)\approx \num{11.1736}\,\G$ and
$F_{\text{round}}(128)=F_{\text{mult}}(128,768,50000)\approx \num{4.9152}\,\G$, we obtain:
\[
T_{\text{dec}}=200,\;\gamma_{\text{clamp}}=0:\quad
F_{\text{infer}}^{\text{DiffLM}}(128)\approx \num{2239.6404}\,\G,
\quad
I_{\text{infer}}^{\text{DiffLM}}=\num{140.5131}.
\]
\[
T_{\text{dec}}=200,\;\gamma_{\text{clamp}}=1:\quad
F_{\text{infer}}^{\text{DiffLM}}(128)\approx \num{3217.7652}\,\G,
\quad
I_{\text{infer}}^{\text{DiffLM}}=\num{201.8799}.
\]
\[
T_{\text{dec}}=2000,\;\gamma_{\text{clamp}}=0:\quad
F_{\text{infer}}^{\text{DiffLM}}(128)\approx \num{22352.1669}\,\G,
\quad
I_{\text{infer}}^{\text{DiffLM}}=\num{1402.356}.
\]
The released decoder clamps at every step by default (the earlier version used $\gamma_{\text{clamp}}{=}0$, 140.5--1402.4$\times$):
\[
T_{\text{dec}}=2000,\;\gamma_{\text{clamp}}=1:\quad
F_{\text{infer}}^{\text{DiffLM}}(128)\approx \num{32177.6517}\,\G,
\quad
I_{\text{infer}}^{\text{DiffLM}}=\num{2018.7986}.
\]

\subsubsection{Summary table (100M tier)}
\label[subsubappendix]{app:diff_family_table_100m}

\begin{table}[H]
\centering
\small
\setlength{\tabcolsep}{4pt}
\begin{tabular}{llrr}
\toprule
Method & Variant
& \multicolumn{1}{c}{$F_{\text{infer}}^{\mathcal{M}}$ ($\G$)}
& \multicolumn{1}{c}{$I_{\text{infer}}^{\mathcal{M}}$} \\
\midrule
AR (100M) & baseline
& \num{15.939} & 1 \\
\midrule
Reviser & $p_{\text{move}}\in\{0.20,0.28,0.33\}$
& \num{19.9709}--\num{23.9006} & \num{1.25}--\num{1.50} \\
\midrule
SEDD & $T_{\text{dec}}{=}32$ (full logits each step)
& \num{514.8424} & \num{32.3008} \\
SEDD & $T_{\text{dec}}{=}2048$ (full logits each step)
& \num{32949.9154} & \num{2067.2497} \\
\midrule
MDLM & $T_{\text{dec}}{=}1000$ (ancestral; full logits each step)
& \num{16088.8259} & \num{1009.3993} \\
MDLM & $T_{\text{dec}}{=}1000$ (released cached sampler)
& \num{1950.0134} & \num{122.3422} \\
\midrule
D3PM & $T_{\text{dec}}{=}1000$ (discrete diffusion; full logits each step)
& \num{16088.8259} & \num{1009.3993} \\
\midrule
Diffusion-LM & $T_{\text{dec}}{=}200,\;\gamma_{\text{clamp}}{=}0$
& \num{2239.6404} & \num{140.5131} \\
Diffusion-LM & $T_{\text{dec}}{=}200,\;\gamma_{\text{clamp}}{=}1$
& \num{3217.7652} & \num{201.8799} \\
Diffusion-LM & $T_{\text{dec}}{=}2000,\;\gamma_{\text{clamp}}{=}0$
& \num{22352.1669} & \num{1402.356} \\
Diffusion-LM & $T_{\text{dec}}{=}2000,\;\gamma_{\text{clamp}}{=}1$ (default)
& \num{32177.6517} & \num{2018.7986} \\
\bottomrule
\end{tabular}
\caption{100M-tier diffusion-family FLOPs multipliers under a matmul-dominant proxy.
(All rows use the shared 100M reference configuration.)}
\end{table}

\subsection{Group 3: Large-Scale Diffusion NAR LMs}

The large backbones used in this subsection (e.g., LLaMA/Qwen families) commonly use
(i) a gated MLP with three linear maps (\texttt{gate\_proj}, \texttt{up\_proj}, \texttt{down\_proj}),
and (ii) grouped-query attention (GQA) where K/V have fewer heads than Q.
Let $h$ be the number of query heads and $h_{\text{kv}}$ the number of K/V heads.
With head dimension $d_{\text{head}}=d/h$, define the effective K/V projection width
$d_{\text{kv}} \triangleq h_{\text{kv}}\,d_{\text{head}}$.
(For standard MHA, $h_{\text{kv}}=h$ so $d_{\text{kv}}=d$.)

For $n$ positions, Q and O are $(n,d)\times(d,d)$, while K and V are $(n,d)\times(d,d_{\text{kv}})$:
\[
F_{\text{proj}}^{\text{GQA}}(n;d,d_{\text{kv}})
\triangleq
2F_{\text{mult}}(n,d,d) + 2F_{\text{mult}}(n,d,d_{\text{kv}}).
\]

A gated MLP uses three matmuls (gate, up, down):
\[
F_{\text{mlp}}^{\text{gated}}(n;d,d_{\text{ff}})
\triangleq
3F_{\text{mult}}(n,d,d_{\text{ff}}).
\]

\[
F_{\text{transformer}}^{\text{full,gated}}(n;d,L,d_{\text{ff}},d_{\text{kv}})
\triangleq
L\Big(
F_{\text{proj}}^{\text{GQA}}(n;d,d_{\text{kv}})
+2F_{\text{mult}}(n,n,d)
+F_{\text{mlp}}^{\text{gated}}(n;d,d_{\text{ff}})
\Big).
\]

\[
F_{\text{transformer}}^{\text{causal,gated}}(n;d,L,d_{\text{ff}},d_{\text{kv}})
\triangleq
L\Big(
F_{\text{proj}}^{\text{GQA}}(n;d,d_{\text{kv}})
+n(n+1)d
+F_{\text{mlp}}^{\text{gated}}(n;d,d_{\text{ff}})
\Big).
\]

\subsubsection{Large-scale reference configurations (native sizes)}
\label[subsubappendix]{app:diff_family_big_configs}

We keep each large model at its native architecture and compare to a size-matched AR baseline of the same scale.
All comparisons use the same output length $n=128$ and the same FLOPs conventions as in \Cref{app:flops_primitives}.

For a decoder-only Transformer, we denote the number of layers by $L$, hidden size by $d$, MLP width by $d_{\text{ff}}$,
and vocabulary size by $|V|$. For attention, let $h$ be the number of query heads and $h_{\text{kv}}$ the number of K/V heads
(Grouped-Query Attention, GQA). With head dimension $d_{\text{head}} \triangleq d/h$, the effective K/V projection width is
$d_{\text{kv}} \triangleq h_{\text{kv}}\,d_{\text{head}}$.
(For standard multi-head attention, $h_{\text{kv}}=h$ so $d_{\text{kv}}=d$.)

\[
\text{Qwen2.5-32B: } L=64,\quad d=5120,\quad d_{\text{ff}}=27648,\quad |V|=152064,\quad
h=40,\quad h_{\text{kv}}=8.
\]
Here $d_{\text{head}}=5120/40=128$ and $d_{\text{kv}}=8\cdot 128=1024$.

\[
\text{Qwen2.5-7B: } L=28,\quad d=3584,\quad d_{\text{ff}}=18944,\quad |V|=152064,\quad
h=28,\quad h_{\text{kv}}=4.
\]
Here $d_{\text{head}}=3584/28=128$ and $d_{\text{kv}}=4\cdot 128=512$.

For LLaDA-8B we use the published LLaDA config (full-attention masked-token predictor):
\[
\text{LLaDA-8B: } L=32,\quad d=4096,\quad d_{\text{ff}}=12288,\quad |V|=126464,\quad
h=32,\quad h_{\text{kv}}=32,
\]
so $d_{\text{head}}=4096/32=128$ and $d_{\text{kv}}=32\cdot 128=4096$ (no GQA reduction).

For the size-matched AR baseline we use LLaMA3-8B (causal decoder with GQA):
\[
\text{LLaMA3-8B (AR): } L=32,\quad d=4096,\quad d_{\text{ff}}=14336,\quad |V|=128256,\quad
h=32,\quad h_{\text{kv}}=8,
\]
so $d_{\text{head}}=128$ and $d_{\text{kv}}=8\cdot 128=1024$.

\subsubsection{AR baselines (size-matched)}
\label[subsubappendix]{app:diff_family_big_ar}

For each backbone, we use a LLaMA/Qwen-style decoder proxy (GQA + gated MLP):
\[
F_{\text{infer}}^{\text{AR}}(n)
\triangleq
F_{\text{transformer}}^{\text{causal,gated}}(n; d,L,d_{\text{ff}},d_{\text{kv}})
+
F_{\text{mult}}(n,d,|V|).
\]

\[
F_{\text{infer}}^{\text{AR (Qwen2.5-32B)}}(128)\approx \num{4099.3869}\,\G.
\]
\[
F_{\text{infer}}^{\text{AR (Qwen2.5-7B)}}(128)\approx \num{906.6536}\,\G.
\]
\[
F_{\text{infer}}^{\text{AR (LLaMA3-8B)}}(128)\approx \num{962.7605}\,\G.
\]

\subsubsection{SDLM-32B-D4 (Sequential Diffusion Language Model)}
\label[subsubappendix]{app:sdlm_32b}

SDLM predicts a fixed block of $D$ tokens per forward pass, then selects the longest high-confidence
prefix (length $k\in\{1,\dots,D\}$) and commits it. KV-cache states for committed tokens are reused.
Here, $D$ is the block size (we use $D=4$), and $\bar{k}\triangleq \mathbb{E}[k]$ is the average committed tokens per pass.

To propose a block of $D$ new tokens given a cached prefix of length $n_{\text{pre}}$, we count:
(i) Q/K/V/O projections + gated MLP on the $D$ new positions, and
(ii) attention matmuls for $D$ queries attending to the cached prefix plus the causal within-block keys.
Let $d_{\text{kv}}$ be the K/V projection width (for GQA; for MHA, $d_{\text{kv}}=d$). We use:
\[
\begin{aligned}
F_{\text{kvblk}}(D,n_{\text{pre}}; d_{\text{ff}}, d_{\text{kv}})
\triangleq {}& L\Big(
\underbrace{2F_{\text{mult}}(D,d,d)+2F_{\text{mult}}(D,d,d_{\text{kv}})}_{\text{Q,O and K,V projections}}
+\underbrace{3F_{\text{mult}}(D,d,d_{\text{ff}})}_{\text{gated MLP}}
\\
&\qquad
+\underbrace{2F_{\text{mult}}(D,n_{\text{pre}},d)+F_{\text{mult}}(D,D{+}1,d)}_{\text{attn (prefix + causal within-block)}}
\Big).
\end{aligned}
\]

With $S\approx \lceil n/\bar{k}\rceil$ passes and prefix length $n_{\text{pre}}\approx (s-1)\bar{k}$, and $\bar{k}=2$:
\[
F_{\text{infer}}^{\text{SDLM}}(n)
\triangleq
\sum_{s=1}^{S}
\Big(
F_{\text{kvblk}}(D,(s-1)\bar{k}; d_{\text{ff}}, d_{\text{kv}})
+
F_{\text{mult}}(D,d,|V|)
\Big).
\]

\[
F_{\text{infer}}^{\text{SDLM-32B-D4}}(128)\approx \num{8198.9416}\,\G,
\quad
I_{\text{infer}}^{\text{SDLM-32B-D4}}=\num{2}.
\]

\subsubsection{Dream-7B (diffusion LLM; Qwen2.5-7B backbone)}
\label[subsubappendix]{app:dream_7b}

Dream performs masked discrete diffusion: starting from a fully masked response, it iteratively denoises
for $T_{\text{dec}}$ steps using a full-attention Transformer denoiser.

Let $m_t$ be the number of masked positions scored at step $t$. For a simple closed form we approximate
a linear schedule: $\sum_{t=1}^{T_{\text{dec}}} m_t \approx n\frac{T_{\text{dec}}+1}{2}$.

\[
F_{\text{infer}}^{\text{Dream}}(n)
\approx
T_{\text{dec}}\,F_{\text{transformer}}^{\text{full,gated}}(n)
+
\frac{T_{\text{dec}}+1}{2}\,F_{\text{vocab}}(n;|V|).
\]

\[
T_{\text{dec}}=10:\quad
F_{\text{infer}}^{\text{Dream-7B}}(128)\approx \num{8768.9307}\,\G,
\quad
I_{\text{infer}}^{\text{Dream-7B}}=\num{9.6718}.
\]
\[
T_{\text{dec}}=20:\quad
F_{\text{infer}}^{\text{Dream-7B}}(128)\approx \num{17502.9816}\,\G,
\quad
I_{\text{infer}}^{\text{Dream-7B}}=\num{19.305}.
\]
Dream's own evaluations use one step per new token ($T_{\text{dec}}=n$):
\[
T_{\text{dec}}=128:\quad
F_{\text{infer}}^{\text{Dream-7B}}(128)\approx \num{111830.7311}\,\G,
\quad
I_{\text{infer}}^{\text{Dream-7B}}=\num{123.3445}.
\]

\subsubsection{LLaDA-8B (diffusion mask predictor; compared to LLaMA3-8B AR)}
\label[subsubappendix]{app:llada_8b}

LLaDA predicts masked tokens with a full-attention Transformer (no causal mask) and runs a reverse masking/denoising
process for $T_{\text{dec}}$ steps. For instruct-style decoding, it can also generate left-to-right in blocks, applying the reverse process inside each block.
Because this configuration uses $h_{\text{kv}}{=}h$ (no GQA reduction), its K/V-side projection terms are larger than in GQA AR baselines, which contributes to the larger compute multiplier.

\[
F_{\text{infer}}^{\text{LLaDA}}(n)
\approx
T_{\text{dec}}\,F_{\text{transformer}}^{\text{full,gated}}(n)
+
\frac{T_{\text{dec}}+1}{2}\,F_{\text{vocab}}(n;|V|).
\]

The released sampler runs the whole sequence at every step, also in its semi-autoregressive block mode (blocks only decide which positions may be unmasked), and allows at most one step per new token ($T_{\text{dec}}\le n$). The earlier version used $T_{\text{dec}}=256$ for $n=128$ (247.5$\times$, outside that range) and a block variant that shortened the sequence to the current block (37.8$\times$), which the sampler does not do. We use the released settings: $T_{\text{dec}}=64$ (four blocks of 32 with 16 steps each) and $T_{\text{dec}}=n=128$:
\[
T_{\text{dec}}=64:\quad
F_{\text{infer}}^{\text{LLaDA-8B}}(128)\approx \num{59604.3482}\,\G,
\quad
I_{\text{infer}}^{\text{LLaDA-8B}}=\num{61.9098}.
\]
\[
T_{\text{dec}}=128:\quad
F_{\text{infer}}^{\text{LLaDA-8B}}(128)\approx \num{119175.5446}\,\G,
\quad
I_{\text{infer}}^{\text{LLaDA-8B}}=\num{123.7852}.
\]

\subsubsection{Summary table (large-scale)}
\label[subsubappendix]{app:diff_family_table_big}

\begin{table}[H]
\centering
\small
\setlength{\tabcolsep}{4pt}
\begin{tabular}{llrr}
\toprule
Method & Variant
& \multicolumn{1}{c}{$F_{\text{infer}}^{\mathcal{M}}$ ($\G$)}
& \multicolumn{1}{c}{$I_{\text{infer}}^{\mathcal{M}}$} \\
\midrule
AR (Qwen2.5-32B) & baseline
& \num{4099.3869} & 1 \\
SDLM-32B-D4 & $D{=}4,\;\bar{k}{=}2$ (no verify)
& \num{8198.9416} & \num{2} \\
\midrule
AR (Qwen2.5-7B) & baseline
& \num{906.6536} & 1 \\
Dream-7B & $T_{\text{dec}}{=}10$ (lin.\ unmask)
& \num{8768.9307} & \num{9.6718} \\
Dream-7B & $T_{\text{dec}}{=}20$ (lin.\ unmask)
& \num{17502.9816} & \num{19.305} \\
Dream-7B & $T_{\text{dec}}{=}128$ (one token per step)
& \num{111830.7311} & \num{123.3445} \\
\midrule
AR (LLaMA3-8B) & baseline
& \num{962.7605} & 1 \\
LLaDA-8B & $T_{\text{dec}}{=}64$ (blocks of 32, 16 steps each)
& \num{59604.3482} & \num{61.9098} \\
LLaDA-8B & $T_{\text{dec}}{=}128$ (one token per step)
& \num{119175.5446} & \num{123.7852} \\
\bottomrule
\end{tabular}
\caption{Large-scale diffusion-family FLOPs multipliers under a matmul-dominant proxy.
Each block uses a size-matched AR baseline (Qwen2.5-32B, Qwen2.5-7B, or LLaMA3-8B).}
\end{table}

\subsection{Group 4. Refinement NAR LMs (MT)}
\subsubsection{Levenshtein Transformer (LevT)}
\citet{gu2019levenshtein} describe the Levenshtein Transformer (LevT), which runs $R$ refinement iterations. Each iteration applies three predictors in sequence: delete $\pi^{\text{del}}$, placeholder-count $\pi^{\text{plh}}$, and token-fill $\pi^{\text{tok}}$. This corresponds to three trunk evaluations per iteration, with optional early exit for the first two:
$f_{\text{del}}=\frac{b_{\text{del}}}{L}$ and $f_{\text{plh}}=\frac{b_{\text{plh}}}{L}$, with $f_{\text{del}},f_{\text{plh}}\in(0,1]$.

We evaluate delete + placeholder predictors at an average length
$n_0 \triangleq 0.8n$, and we evaluate the token-fill predictor at length $n$ (after insertion returns to the target length). For the token-fill vocabulary head, we assume only placeholder positions require vocab logits, with $m \triangleq 0.4n$ denoting the estimated number of placeholder tokens to fill at each refinement iteration.
This LevT accounting is a decoder-side iterative-edit proxy (matching the LevT edit cycle) rather than a full source-conditioned MT accounting with separate encoder/cross-attention terms.

LevT runs bidirectional/refinement-style passes, so we use full attention in the trunk terms:
\[
\begin{aligned}
F_{\text{iter}}^{\text{LevT}}(n)
\;\approx\;
&\underbrace{f_{\text{del}}\,F_{\text{transformer}}^{\text{full}}(n_0)}_{\text{delete trunk}}
+\underbrace{f_{\text{plh}}\,F_{\text{transformer}}^{\text{full}}(n_0)}_{\text{placeholder trunk}}
+\underbrace{F_{\text{transformer}}^{\text{full}}(n)}_{\text{token trunk}}\\
&+\underbrace{F_{\text{mult}}(n_0,d,2)}_{\text{delete head}}
+\underbrace{F_{\text{mult}}(n_0+1,2d,K_{\max}+1)}_{\text{placeholder-count head}}
+\underbrace{F_{\text{vocab}}(m)}_{\text{vocab projection on placeholders}},
\end{aligned}
\]
where $n_0=0.8n$ and $m=0.4n$. These are average-case analytical surrogates and need not be integers. Total decoding cost is
\[
F_{\text{infer}}^{\text{LevT}}(n)\approx R\cdot F_{\text{iter}}^{\text{LevT}}(n).
\]

Using the shared reference configuration with $n=128$, $n_0=102.4$, and $m=51.2$:
\[
F_{\text{transformer}}^{\text{full}}(102.4)\approx \num{8.8906}\,\G,\quad
F_{\text{transformer}}^{\text{full}}(128)\approx \num{11.1736}\,\G,
\]
and head costs
\[
F_{\text{mult}}(102.4,768,2)\approx \num{0.1573}\,\M,\quad
F_{\text{mult}}(103.4,1536,17)\approx \num{2.7}\,\M,\quad
F_{\text{vocab}}(51.2)\approx \num{1.9661}\,\G.
\]

For reference, the earlier version's settings (fixed $R\in\{5,10\}$, with and without half-depth early exit) give the following. Set $f_{\text{del}}=f_{\text{plh}}=1$. Then:
\[
F_{\text{iter,noEE}}^{\text{LevT}}(128)\approx \num{30.9237}\,\G.
\]
With $R=5$ and $R=10$:
\[
F_{\text{infer,noEE}}^{\text{LevT}}(128,R=5)\approx \num{154.6186}\,\G,\qquad
F_{\text{infer,noEE}}^{\text{LevT}}(128,R=10)\approx \num{309.2373}\,\G.
\]

Set $f_{\text{del}}=f_{\text{plh}}=0.5$. Then:
\[
F_{\text{iter,EE}}^{\text{LevT}}(128)\approx \num{22.0331}\,\G.
\]
With $R=5$ and $R=10$:
\[
F_{\text{infer,EE}}^{\text{LevT}}(128,R=5)\approx \num{110.1657}\,\G,\qquad
F_{\text{infer,EE}}^{\text{LevT}}(128,R=10)\approx \num{220.3315}\,\G.
\]

Using $F_{\text{infer}}^{\text{AR}}(128)\approx \num{15.939}\,\G$:
\[
I_{\text{infer,noEE}}^{\text{LevT}}(R=5)=\num{9.7006},
\quad
I_{\text{infer,noEE}}^{\text{LevT}}(R=10)=\num{19.4013},
\]
\[
I_{\text{infer,EE}}^{\text{LevT}}(R=5)=\num{6.9117},
\quad
I_{\text{infer,EE}}^{\text{LevT}}(R=10)=\num{13.8234}.
\]
The released checkpoint has no early-exit heads, and its adaptive stop ends decoding after $R=3.27$ iterations on average with caps of 5 and 10 (\Cref{app:measured_flops}). With $f_{\text{del}}=f_{\text{plh}}=1$:
\[
F_{\text{infer}}^{\text{LevT}}(128,R=3.27)\approx \num{101.1206}\,\G,
\qquad
I_{\text{infer}}^{\text{LevT}}=\num{6.3442}.
\]

\subsubsection{Insertion-based Generation: Insertion Transformer (Stern et al., 2019)}
\citet{stern2019insertion} propose the Insertion Transformer (InsT), which generates a sequence by iteratively inserting tokens into slots (gaps). At iteration $t$, the current partial sequence has length $n_t$. The model runs a Transformer pass on the current sequence and predicts, for each slot, either a vocabulary token to insert or a special \texttt{no-insert} action.

Insertion Transformer uses bidirectional passes over the current partial sequence, so we use full attention:
\[
F_{\text{iter}}^{\text{InsT}}(n_t)
\triangleq
F_{\text{transformer}}^{\text{full}}(n_t)
+
F_{\text{vocab}}(n_t+1),
\]
where $n_t+1$ slots are scored. (The extra \texttt{no-insert} class changes $|V_c|$ by $+1$, which is negligible at $|V_c|=50\text{k}$.)
AR baseline: $F_{\text{infer}}^{\text{AR}}(128)\approx \num{15.939}\,\G$.
Take $n_t\in\{1,2,4,8,16,32,64,128\}$. Then:
\[
\sum_t F_{\text{transformer}}^{\text{full}}(n_t)\approx \num{22.061}\,\G,\qquad
\sum_t F_{\text{vocab}}(n_t+1)\approx \num{10.0992}\,\G,
\]
so
\[
F_{\text{infer,bal}}^{\text{InsT}}(128)\approx \num{32.1602}\,\G,\qquad
I_{\text{infer,bal}}^{\text{InsT}}=\num{2.0177}.
\]

Take $n_t\in\{1,2,\dots,128\}$. Then:
\[
\sum_t F_{\text{transformer}}^{\text{full}}(n_t)\approx \num{714.2568}\,\G,\qquad
\sum_t F_{\text{vocab}}(n_t+1)\approx \num{321.9456}\,\G,
\]
so
\[
F_{\text{infer,ser}}^{\text{InsT}}(128)\approx \num{1036.2024}\,\G,\qquad
I_{\text{infer,ser}}^{\text{InsT}}=\num{65.0105}.
\]

\subsubsection{One-shot NAT (fertility / length latents): Non-Autoregressive NMT (Gu et al., 2018)}
\citet{gu2018nonautoregressive} propose one-shot non-autoregressive translation (NAT), which predicts all target tokens in parallel by using discrete fertility/length latents and a single encoder--decoder pass.

For this numerical instantiation we use $n_{\text{src}}=n=128$. Let $K_{\text{fert}}$ denote the number of discrete fertility classes (the maximum fertility bucketed by the model), so each source token predicts a fertility $f_i \in \{0,1,\dots,K_{\text{fert}}-1\}$; here we set $K_{\text{fert}}=50$.

One-shot NAT runs a single encoder--decoder evaluation:
\[
F_{\text{infer}}^{\text{NAT}}(n,n_{\text{src}})
\triangleq
F_{\text{enc}}(n_{\text{src}})
+F_{\text{mult}}(n_{\text{src}}, d, K_{\text{fert}})
+\Big(F_{\text{transformer}}^{\text{full}}(n)+F_{\text{xattn}}(n,n_{\text{src}})\Big)
+F_{\text{vocab}}(n).
\]

With $n=n_{\text{src}}=128$:
\[
F_{\text{xattn}}(128,128)\approx \num{3.9259}\,\G,
\qquad
F_{\text{fert}}(128)=F_{\text{mult}}(128,768,50)\approx \num{4.9152}\,\M.
\]
Therefore $F_{\text{infer}}^{\text{NAT}}(128,128)\approx \num{31.1932}\,\G$.
Because NAT has an encoder, we compare it with an encoder--decoder AR model of the same size; with a KV cache, each target position runs its cross-attention once:
\[
F_{\text{infer}}^{\text{AR,MT}}(n,n_{\text{src}})
\triangleq
F_{\text{enc}}(n_{\text{src}})
+F_{\text{transformer}}^{\text{causal}}(n)
+F_{\text{xattn}}(n,n_{\text{src}})
+F_{\text{vocab}}(n),
\]
$F_{\text{infer}}^{\text{AR,MT}}(128,128)\approx \num{31.0385}\,\G$, so $I_{\text{infer}}^{\text{NAT}}=\num[round-precision=3]{1.005}$. (The earlier version divided by the decoder-only AR, $\num{15.939}\,\G$, giving \num{1.957}.)

\subsubsection{Iterative Mask Refinement: Mask-Predict / CMLM (Ghazvininejad et al., 2019)}
\citet{ghazvininejad2019maskpredict} start Mask-Predict from a fully-masked target and perform $T_{\text{mp}}$ refinement cycles, each time masking low-confidence tokens and re-predicting them in parallel.

We instantiate $T_{\text{mp}}=10$ and the common linear decay schedule:
\[
m_0=n,\qquad
m_t=\left\lfloor n\cdot\frac{T_{\text{mp}}-t}{T_{\text{mp}}}\right\rfloor \ \ (t=1,\dots,T_{\text{mp}}-1).
\]
For $n=128$, $T_{\text{mp}}=10$:
\[
(m_0,\dots,m_9)=(128,115,102,89,76,64,51,38,25,12),
\qquad
\sum_{t=0}^{9} m_t = 700.
\]

At each iteration we pay decoder trunk + cross-attention + vocab projection on only the masked tokens:
\[
F_{\text{iter}}^{\text{CMLM}}(t, n, n_{\text{src}})
=
F_{\text{transformer}}^{\text{full}}(n)
+
F_{\text{xattn}}(n,n_{\text{src}})
+
F_{\text{vocab}}(m_t).
\]
Encoder cost is paid once:
$F_{\text{enc}}(n_{\text{src}})=F_{\text{transformer}}^{\text{full}}(n_{\text{src}})$.

With $n_{\text{src}}=n=128$:
\[
F_{\text{transformer}}^{\text{full}}(128)+F_{\text{xattn}}(128,128)\approx \num{15.0995}\,\G.
\]
Masked-only vocab projection across all iterations:
\[
\sum_{t=0}^{T_{\text{mp}}-1} F_{\text{vocab}}(m_t)
\approx \num{26.88}\,\G.
\]
Mask-Predict decodes $\ell$ length candidates and keeps the most probable; the encoder is shared, and each candidate pays all iterations (we approximate every candidate at length $n$):
\[
F_{\text{infer}}^{\text{CMLM}}(128,128;\ell)
=
F_{\text{enc}}(128)+\ell\sum_{t=0}^{T_{\text{mp}}-1}F_{\text{iter}}^{\text{CMLM}}(t,128,128).
\]
Against the encoder--decoder AR baseline $F_{\text{infer}}^{\text{AR,MT}}(128,128)$ (NAT above):
\[
\ell=1:\ \num{189.0486}\,\G,\ I_{\text{infer}}^{\text{CMLM}}=\num{6.0908};
\qquad
\ell=5 \text{ (the authors' default)}:\ \num{900.5483}\,\G,\ I_{\text{infer}}^{\text{CMLM}}=\num{29.0139}.
\]
(The earlier version used $\ell=1$ and the decoder-only AR baseline, giving \num{11.8607}.)

\section{Example Trajectories}
\label[appendix]{sec:appendix_ranked_examples}
\textbf{Step-by-step replays.} All 70 trajectories on the shortlist described below, including the examples shown here, can be replayed one action at a time at \url{https://sean-diab.github.io/reviser_trajectories.html}.

The examples below were selected by hand, not sampled at random; \Cref{sec:results_cont,sec:results_edit} give the measured quality. For each continuation model, an automatic filter built a pool of 100 outputs that stopped by themselves, have 50--180 new tokens, are not repetitive, have a GPT-2 perplexity at or below the model's median, contain no personal information and use the cursor visibly (for the 100M pair, the AR text on the same prompt also had to pass a perplexity filter). For each edit model there are two pools: 100 items whose output changes the input and ends closer to the reference than the input was, and 65 items made of the 50 outputs that match the reference exactly and change the input most, plus 15 from the less common edit categories (grammar, fluency, clarity, coherence, paraphrase, style), most of them a few tokens from the reference, since exact matches outside neutralization are rare. An LLM reading agent picked ten candidates from each pool, and we chose the examples shown here from those. \textbf{Notation.} Prompt or draft text is not repeated in the actions. Consecutive \Insert{} actions are merged into one typed span (\textcolor{FigTyrianPurple}{\texttt{\textquotedbl{}~the~set\textquotedbl{}}}), consecutive \Move{} actions into one bracket that lists each move (\textcolor{FigBlue}{\texttt{[-{}16 -{}4 -{}1]}} jumps 21 tokens back), \textcolor{FigRed}{\texttt{[Del]}} is one \Delete{} (a backspace) and \textcolor{FigRed}{\texttt{[Stop]}} ends the trajectory. In the edit examples, words removed from the input are red and words the output adds are green.

\subsection{Continuation}
\subsubsection*{Reviser 100M (run-and-jump + \Delete{} + RL) and AR 100M}
\textbf{Reviser 100M} (prompt in blue; 116 new tokens, 142 actions: 120 inserts, 17 moves, 4 deletes).\par
{\ttfamily\footnotesize\raggedright \textcolor{blue}{The Company applies the Quality Management System described in their Company Documented Information to provide quality assurance to their clients and continual improvement. The Documented Information has been developed in accordance with} the set of standards and the quality management system provided by the Company.\textbackslash{}n\newline If one is not satisfied from the quality, it becomes the responsibility of the customer to be responsible for the accuracy of the information offered by the company. We take excellent care in the preparation of such documents. If we believe you have not addressed the problem, we will give you the right to keep our website updated with accurate, accurate information about the company.\textbackslash{}n\newline We hope that you can make sure that you are satisfied.\textbackslash{}n\newline If you wish to reach us, please contact our Customer Care office for guidance.\par}
\vspace{0.3em}\textbf{AR 100M}, same prompt, cut to the same length (116 of 220 new tokens).\par
{\ttfamily\footnotesize\raggedright \textcolor{blue}{The Company applies the Quality Management System described in their Company Documented Information to provide quality assurance to their clients and continual improvement. The Documented Information has been developed in accordance with} the Company\textquotesingle{}s Quality Management System. Each document is provided as a service document.\textbackslash{}n\newline The Quality Management System is designed to provide a quality management system for both customers and suppliers to ensure that all necessary materials are used in the manufacture, distribution, delivery, and delivery of goods and services. The Quality Management System allows for a system that offers timely, timely service and helps customers to ensure that necessary materials are stored properly.\textbackslash{}n\newline The Quality Management System is a product that enables the Company to increase its production capacity with increased efficiency in order to provide an optimal product range. Quality Management System is\par}
\vspace{0.3em}\textbf{What happens.} The model writes the closing request as a frame, `If, please contact our Customer Care.', then steps back to add `office for guidance' and, nine tokens further back, the missing clause `you wish to reach us'. Later it jumps 39 tokens back to the start of its text and finishes the prompt's `developed in accordance with': it types `the set and', then `standards', steps back, types `standards' again, removes the repeat at once with \Delete{} and writes `of', giving `the set of standards and'. The repeat `accurate, accurate information' stays.\par
\vspace{0.3em}\textbf{Actions} after the prompt.\par
{\ttfamily\footnotesize\raggedright\setlength{\parindent}{0pt}\textcolor{FigTyrianPurple}{\textquotedbl{}~the quality management system provided by the Company.\textbackslash{}nIf one is not satisfied from the quality, it becomes the responsibility of the. We take excellent care that you can make sure that you are satisfied.\textbackslash{}nIf, please contact our.\textquotedbl{}} \textcolor{FigRed}{[Del]} \textcolor{FigTyrianPurple}{\textquotedbl{}~Customer Care.\textquotedbl{}} \textcolor{FigBlue}{[-{}1]} \textcolor{FigTyrianPurple}{\textquotedbl{}~office for guidance\textquotedbl{}} \textcolor{FigBlue}{[-{}8 -{}1]} \textcolor{FigTyrianPurple}{\textquotedbl{}~you wish to reach us\textquotedbl{}} \textcolor{FigBlue}{[-{}16 -{}1]} \textcolor{FigTyrianPurple}{\textquotedbl{}~in the preparation of such documents. If we believe you have not the\textquotedbl{}} \textcolor{FigRed}{[Del]} \textcolor{FigTyrianPurple}{\textquotedbl{}~addressed the problem, we will give to\textquotedbl{}} \textcolor{FigRed}{[Del]} \textcolor{FigTyrianPurple}{\textquotedbl{}~you the right to keep accurate, accurate information about the.\textbackslash{}nWe hope\textquotedbl{}} \textcolor{FigBlue}{[-{}4]} \textcolor{FigTyrianPurple}{\textquotedbl{}~company\textquotedbl{}} \textcolor{FigBlue}{[-{}4 -{}2 -{}1]} \textcolor{FigTyrianPurple}{\textquotedbl{}~our website updated with\textquotedbl{}} \textcolor{FigBlue}{[-{}32 -{}2]} \textcolor{FigTyrianPurple}{\textquotedbl{}~customer to be responsible for the accuracy of the information offered by the company\textquotedbl{}} \textcolor{FigBlue}{[-{}32 -{}4 -{}2 -{}1]} \textcolor{FigTyrianPurple}{\textquotedbl{}~the set and\textquotedbl{}} \textcolor{FigBlue}{[-{}1]} \textcolor{FigTyrianPurple}{\textquotedbl{}~standards\textquotedbl{}} \textcolor{FigBlue}{[-{}1]} \textcolor{FigTyrianPurple}{\textquotedbl{}~standards\textquotedbl{}} \textcolor{FigRed}{[Del]} \textcolor{FigTyrianPurple}{\textquotedbl{}~of\textquotedbl{}} \textcolor{FigRed}{[Stop]}\par}
\vspace{0.8em}
\subsubsection*{Reviser 300M (run-and-jump + \Delete{} + RL)}
\textbf{Reviser 300M} (prompt in blue; 126 new tokens, 157 actions: 129 inserts, 24 moves, 3 deletes).\par
{\ttfamily\footnotesize\raggedright \textcolor{blue}{We were in a posh part of Kent (no not Dover!!) somewhere near Ashford on Thursday night just gone. We had been hired to ensure a well known (}and still very popular) restaurant in the region, where the wine was served. It was a great day out, but it was still a little cold when we left, so we made the transition to a nice bar for a drink. We were soon treated to a fantastic wine tasting and a lovely drink in the evening.\textbackslash{}n\newline As expected, there were great food, great drinks and good wine. The staff and service were extremely friendly and welcoming, and there was a nice mix of food and wine. The atmosphere was warm and welcoming, with a few of the staff making some excellent connections.\textbackslash{}n\newline All in all a great experience.\par}
\vspace{0.3em}\textbf{What happens.} The model first writes a skeleton of the review, `restaurant. As expected, there were great food, great The staff and staff', removes the repeated `staff' at once and continues `service were extremely friendly and welcoming', then writes the closing line `All in all a great experience.' It then jumps back to fill what it skipped: `great' becomes `great drinks and good wine', `and there was a nice mix of food and wine' goes after `welcoming,', and a 54-token jump back to `restaurant' adds the whole first paragraph (`in the region, where the wine was served. It was a great day out, but \ldots'), itself filled in from the middle out. The ungrammatical `there were great food' stays.\par
\vspace{0.3em}\textbf{Actions} after the prompt.\par
{\ttfamily\footnotesize\raggedright\setlength{\parindent}{0pt}\textcolor{FigTyrianPurple}{\textquotedbl{}and still very popular) restaurant.\textbackslash{}nAs expected, there.\textquotedbl{}} \textcolor{FigBlue}{[-{}1]} \textcolor{FigTyrianPurple}{\textquotedbl{}~were great food, great The staff and staff\textquotedbl{}} \textcolor{FigRed}{[Del]} \textcolor{FigTyrianPurple}{\textquotedbl{}~service were extremely and\textquotedbl{}} \textcolor{FigRed}{[Del]} \textcolor{FigTyrianPurple}{\textquotedbl{}~friendly and welcoming,. The atmosphere was,\textquotedbl{}} \textcolor{FigRed}{[Del]} \textcolor{FigTyrianPurple}{\textquotedbl{}~warm and welcoming, with a few of the staff making\textbackslash{}nAll in all a great experience\textquotedbl{}} \textcolor{FigBlue}{[-{}16 -{}4 -{}2]} \textcolor{FigTyrianPurple}{\textquotedbl{}~and there was a nice mix of food and wine\textquotedbl{}} \textcolor{FigBlue}{[-{}16 -{}4]} \textcolor{FigTyrianPurple}{\textquotedbl{}~drinks and.\textquotedbl{}} \textcolor{FigBlue}{[-{}1]} \textcolor{FigTyrianPurple}{\textquotedbl{}~good wine\textquotedbl{}} \textcolor{FigBlue}{[+32 +4]} \textcolor{FigTyrianPurple}{\textquotedbl{}~some excellent.\textquotedbl{}} \textcolor{FigBlue}{[-{}1]} \textcolor{FigTyrianPurple}{\textquotedbl{}~connections\textquotedbl{}} \textcolor{FigBlue}{[-{}32 -{}16 -{}4 -{}2]} \textcolor{FigTyrianPurple}{\textquotedbl{}~in the region, was a great day out, but\textquotedbl{}} \textcolor{FigBlue}{[-{}4 -{}2 -{}1]} \textcolor{FigTyrianPurple}{\textquotedbl{}~where the wine was served. It\textquotedbl{}} \textcolor{FigBlue}{[+4 +2 +1]} \textcolor{FigTyrianPurple}{\textquotedbl{}~it was still a little cold we left\textquotedbl{}} \textcolor{FigBlue}{[-{}2]} \textcolor{FigTyrianPurple}{\textquotedbl{}~when\textquotedbl{}} \textcolor{FigBlue}{[+2]} \textcolor{FigTyrianPurple}{\textquotedbl{}, so we made the transition to a nice bar for a drink. We were soon treated to a fantastic wine tasting and a drink\textquotedbl{}} \textcolor{FigBlue}{[-{}1]} \textcolor{FigTyrianPurple}{\textquotedbl{}~lovely\textquotedbl{}} \textcolor{FigBlue}{[+1]} \textcolor{FigTyrianPurple}{\textquotedbl{}~in the evening\textquotedbl{}} \textcolor{FigRed}{[Stop]}\par}
\vspace{0.8em}
\subsubsection*{Released Reviser 100M}
\textbf{Released Reviser 100M} (prompt in blue; 77 new tokens, 114 actions: 77 inserts, 36 moves, 0 deletes).\par
{\ttfamily\footnotesize\raggedright \textcolor{blue}{This is entertaining made easy and it\textquoteright{}s extra tasty.\textbackslash{}n\newline Heat the oil in a large pan or wok.\textbackslash{}n\newline Add the curry paste, squash and lemong}rass, onion, pepper, garlic, salt and salt. add in the onion and garlic and add the pasta and pepper to taste and then add in the garlic powder.\textbackslash{}n\newline Let it boil and the pasta and season it with salt and pepper, until just combined to taste.\textbackslash{}n\newline This is one of the better meals I\textquoteright{}ve had, but the chicken is so hot!\par}
\vspace{0.3em}\textbf{What happens.} The prompt stops in the middle of `lemongrass'. The released model's first actions type the final `!' and then `is hot'; it builds the text outward from there with 36 moves, several of which undo the one before (\texttt{[-{}1 +1 -{}1]}), and types the missing `rass' with its last two inserts, 112 actions after the start. The recipe itself makes no sense (`salt and salt', pasta in a curry, a chicken never mentioned before).\par
\vspace{0.3em}\textbf{Actions} after the prompt.\par
{\ttfamily\footnotesize\raggedright\setlength{\parindent}{0pt}\textcolor{FigTyrianPurple}{\textquotedbl{}!\textquotedbl{}} \textcolor{FigBlue}{[-{}1 +1 -{}1]} \textcolor{FigTyrianPurple}{\textquotedbl{}~is\textquotedbl{}} \textcolor{FigBlue}{[-{}1 +1]} \textcolor{FigTyrianPurple}{\textquotedbl{}~hot\textquotedbl{}} \textcolor{FigBlue}{[-{}2]} \textcolor{FigTyrianPurple}{\textquotedbl{}.\textbackslash{}n\textquotedbl{}} \textcolor{FigBlue}{[+1]} \textcolor{FigTyrianPurple}{\textquotedbl{}~so\textquotedbl{}} \textcolor{FigBlue}{[-{}4 +4 -{}4]} \textcolor{FigTyrianPurple}{\textquotedbl{},\textquotedbl{}} \textcolor{FigBlue}{[-{}1 +2 +1]} \textcolor{FigTyrianPurple}{\textquotedbl{}This\textquotedbl{}} \textcolor{FigBlue}{[-{}4 +1]} \textcolor{FigTyrianPurple}{\textquotedbl{}~taste\textquotedbl{}} \textcolor{FigBlue}{[-{}2 +2 +4]} \textcolor{FigTyrianPurple}{\textquotedbl{}~one,\textquotedbl{}} \textcolor{FigBlue}{[-{}8]} \textcolor{FigTyrianPurple}{\textquotedbl{}~onion and\textquotedbl{}} \textcolor{FigBlue}{[-{}2]} \textcolor{FigTyrianPurple}{\textquotedbl{}~and\textquotedbl{}} \textcolor{FigBlue}{[+2 +4 +2 -{}8]} \textcolor{FigTyrianPurple}{\textquotedbl{}~salt. add in the\textquotedbl{}} \textcolor{FigBlue}{[-{}4 -{}2]} \textcolor{FigTyrianPurple}{\textquotedbl{}, pepper, garlic, salt\textquotedbl{}} \textcolor{FigBlue}{[+16]} \textcolor{FigTyrianPurple}{\textquotedbl{}~but the is\textquotedbl{}} \textcolor{FigBlue}{[-{}1]} \textcolor{FigTyrianPurple}{\textquotedbl{}~chicken\textquotedbl{}} \textcolor{FigBlue}{[-{}4]} \textcolor{FigTyrianPurple}{\textquotedbl{}~of the better meals I\textquoteright{}ve had\textquotedbl{}} \textcolor{FigBlue}{[-{}16]} \textcolor{FigTyrianPurple}{\textquotedbl{}~garlic and add the pasta and pepper to taste and then add in the garlic powder.\textbackslash{}nLet it boil and the pasta and season it with salt and pepper\textquotedbl{}} \textcolor{FigBlue}{[+1]} \textcolor{FigTyrianPurple}{\textquotedbl{}~until just combined to\textquotedbl{}} \textcolor{FigBlue}{[+16 -{}64 -{}2]} \textcolor{FigTyrianPurple}{\textquotedbl{}, onion\textquotedbl{}} \textcolor{FigBlue}{[-{}2]} \textcolor{FigTyrianPurple}{\textquotedbl{}rass\textquotedbl{}} \textcolor{FigBlue}{[-{}2]} \textcolor{FigRed}{[Stop]}\par}
\vspace{0.8em}

\subsection{Editing}
\subsubsection*{Reviser-Edit 100M, IteraTeR (clarity)}
\textbf{Instruction:} Make this text clearer.\par
\vspace{0.2em}\textbf{Input}.\par
{\ttfamily\footnotesize\raggedright By 2008\textcolor{TokRed}{\textvisiblespace{}} the factory at Palmela will be \textcolor{TokRed}{only} producing the multi-{}purpose vehicle Sharan and the Eos models.\par}
\vspace{0.2em}\textbf{Reviser-Edit 100M} (SARI 100.0; 11 actions: 1 insert, 7 moves, 2 deletes).\par
{\ttfamily\footnotesize\raggedright By 2008 the factory at Palmela will \textcolor{TokGreen}{only} be producing the multi-{}purpose vehicle Sharan and the Eos models.\par}
\vspace{0.2em}\textbf{AR rewriter} (SARI 74.5).\par
{\ttfamily\footnotesize\raggedright By 2008 the factory at Palmela will be producing the multi-{}purpose vehicle Sharan and the Eos models.\par}
\vspace{0.2em}\textbf{Reference}.\par
{\ttfamily\footnotesize\raggedright By 2008 the factory at Palmela will \textcolor{TokGreen}{only} be producing the multi-{}purpose vehicle Sharan and the Eos models.\par}
\vspace{0.3em}\textbf{What happens.} The model jumps 14 tokens back from the end, deletes `only', steps back one token and types `only' before `be' (a word move, done as a delete and an insert), then jumps to the start and removes the doubled space after `2008'. The output matches the reference exactly; the AR rewriter drops `only', which changes the meaning.\par
\vspace{0.3em}\textbf{Actions} on the draft.\par
{\ttfamily\footnotesize\raggedright\setlength{\parindent}{0pt}\textcolor{FigBlue}{[-{}8 -{}4 -{}2]} \textcolor{FigRed}{[Del]} \textcolor{FigBlue}{[-{}1]} \textcolor{FigTyrianPurple}{\textquotedbl{}~only\textquotedbl{}} \textcolor{FigBlue}{[-{}4 -{}2 -{}1]} \textcolor{FigRed}{[Del]} \textcolor{FigRed}{[Stop]}\par}
\vspace{0.8em}
\subsubsection*{Reviser-Edit 100M, CoEdIT (grammar)}
\textbf{Instruction:} Fix grammar in this sentence.\par
\vspace{0.2em}\textbf{Input}.\par
{\ttfamily\footnotesize\raggedright Fatty acids found in breast milk \textcolor{TokRed}{has} been found to \textcolor{TokRed}{being} essential for infant\textquotesingle{}s early cognitive development.\par}
\vspace{0.2em}\textbf{Reviser-Edit 100M} (SARI 88.6; 11 actions: 3 inserts, 5 moves, 2 deletes).\par
{\ttfamily\footnotesize\raggedright Fatty acids found in breast milk \textcolor{TokGreen}{have} been found to \textcolor{TokGreen}{be} essential for \textcolor{TokGreen}{the} infant\textquotesingle{}s early cognitive development.\par}
\vspace{0.2em}\textbf{AR rewriter} (SARI 77.2).\par
{\ttfamily\footnotesize\raggedright \textcolor{TokGreen}{The fat} acids found in breast milk \textcolor{TokGreen}{have} been found to \textcolor{TokGreen}{be} essential for infant\textquotesingle{}s early cognitive development.\par}
\vspace{0.2em}\textbf{Reference}.\par
{\ttfamily\footnotesize\raggedright Fatty acids found in breast milk \textcolor{TokGreen}{have} been found to \textcolor{TokGreen}{be} essential for \textcolor{TokGreen}{an} infant\textquotesingle{}s early cognitive development.\par}
\vspace{0.3em}\textbf{What happens.} Three grammar fixes, made from the end of the sentence back to its start: the model jumps six tokens back and types `the' before `infant's', steps back three tokens and replaces `being' with `be' (one \Delete{}, one typed word), then four tokens further back replaces `has' with `have'. The output is correct and differs from the reference in one word (`the infant's' where the reference has `an infant's'). The AR rewriter also fixes `has' and `being', but turns `Fatty acids' into `The fat acids' and leaves `for infant's'.\par
\vspace{0.3em}\textbf{Actions} on the draft.\par
{\ttfamily\footnotesize\raggedright\setlength{\parindent}{0pt}\textcolor{FigBlue}{[-{}4 -{}2]} \textcolor{FigTyrianPurple}{\textquotedbl{}~the\textquotedbl{}} \textcolor{FigBlue}{[-{}2 -{}1]} \textcolor{FigRed}{[Del]} \textcolor{FigTyrianPurple}{\textquotedbl{}~be\textquotedbl{}} \textcolor{FigBlue}{[-{}4]} \textcolor{FigRed}{[Del]} \textcolor{FigTyrianPurple}{\textquotedbl{}~have\textquotedbl{}} \textcolor{FigRed}{[Stop]}\par}
\vspace{0.8em}
\subsubsection*{Reviser-Edit 300M, CoEdIT (grammar), example 1}
\textbf{Instruction:} Fix grammar.\par
\vspace{0.2em}\textbf{Input}.\par
{\ttfamily\footnotesize\raggedright It \textcolor{TokRed}{offered} recommendations for various stores, fashion matching, favorite competitive products, special \textcolor{TokRed}{sale} of new arrivals, editor\textquotesingle{}s \textcolor{TokRed}{choice}, and more.\par}
\vspace{0.2em}\textbf{Reviser-Edit 300M} (SARI 100.0; 14 actions: 3 inserts, 7 moves, 3 deletes).\par
{\ttfamily\footnotesize\raggedright It \textcolor{TokGreen}{offers} recommendations for various stores, fashion matching, favorite competitive products, special \textcolor{TokGreen}{sales} of new arrivals, editor\textquotesingle{}s \textcolor{TokGreen}{choices}, and more.\par}
\vspace{0.2em}\textbf{AR rewriter} (SARI 90.2).\par
{\ttfamily\footnotesize\raggedright It \textcolor{TokGreen}{offers} recommendations for various stores, fashion matching, favorite competitive products, special \textcolor{TokGreen}{sales} of new arrivals, editor\textquotesingle{}s choice, and more.\par}
\vspace{0.2em}\textbf{Reference}.\par
{\ttfamily\footnotesize\raggedright It \textcolor{TokGreen}{offers} recommendations for various stores, fashion matching, favorite competitive products, special \textcolor{TokGreen}{sales} of new arrivals, editor\textquotesingle{}s \textcolor{TokGreen}{choices}, and more.\par}
\vspace{0.3em}\textbf{What happens.} Three fixes from the end of the sentence back to its start, each a jump, one \Delete{} and one typed word: `choice'$\to$`choices', then seven tokens back `sale'$\to$`sales', then a 14-token jump to the second word for `offered'$\to$`offers'. The output matches the reference exactly; the AR rewriter makes the last two fixes but keeps `editor's choice'.\par
\vspace{0.3em}\textbf{Actions} on the draft.\par
{\ttfamily\footnotesize\raggedright\setlength{\parindent}{0pt}\textcolor{FigBlue}{[-{}4]} \textcolor{FigRed}{[Del]} \textcolor{FigTyrianPurple}{\textquotedbl{}~choices\textquotedbl{}} \textcolor{FigBlue}{[-{}4 -{}2 -{}1]} \textcolor{FigRed}{[Del]} \textcolor{FigTyrianPurple}{\textquotedbl{}~sales\textquotedbl{}} \textcolor{FigBlue}{[-{}8 -{}4 -{}2]} \textcolor{FigRed}{[Del]} \textcolor{FigTyrianPurple}{\textquotedbl{}~offers\textquotedbl{}} \textcolor{FigRed}{[Stop]}\par}
\vspace{0.8em}
\subsubsection*{Reviser-Edit 300M, CoEdIT (grammar), example 2}
\textbf{Instruction:} Fix grammatical mistakes in this sentence.\par
\vspace{0.2em}\textbf{Input}.\par
{\ttfamily\footnotesize\raggedright I\textquotesingle{}m sitting here with a stupid, puzzled \textcolor{TokRed}{looks} on my \textcolor{TokRed}{faces} \textcolor{TokRed}{tried} to figure out how we \textcolor{TokRed}{letting} Mo Williams score 52 \textcolor{TokRed}{point}...\par}
\vspace{0.2em}\textbf{Reviser-Edit 300M} (SARI 98.3; 17 actions: 5 inserts, 6 moves, 5 deletes).\par
{\ttfamily\footnotesize\raggedright I\textquotesingle{}m sitting here with a stupid, puzzled \textcolor{TokGreen}{look} on my \textcolor{TokGreen}{face} \textcolor{TokGreen}{trying} to figure out how we \textcolor{TokGreen}{let} Mo Williams score 52 \textcolor{TokGreen}{points}...\par}
\vspace{0.2em}\textbf{AR rewriter} (SARI 98.3).\par
{\ttfamily\footnotesize\raggedright I\textquotesingle{}m sitting here with a stupid, puzzled \textcolor{TokGreen}{look} on my \textcolor{TokGreen}{face} \textcolor{TokGreen}{trying} to figure out how we \textcolor{TokGreen}{let} Mo Williams score 52 \textcolor{TokGreen}{points}...\par}
\vspace{0.2em}\textbf{Reference}.\par
{\ttfamily\footnotesize\raggedright I\textquotesingle{}m sitting here with a stupid, puzzled \textcolor{TokGreen}{look} on my \textcolor{TokGreen}{face} \textcolor{TokGreen}{trying} to figure out how we \textcolor{TokGreen}{let} Mo Williams score 52 \textcolor{TokGreen}{points}.\par}
\vspace{0.3em}\textbf{What happens.} Four fixes from the end back to the start: `point'$\to$`points'; five tokens back, `letting'$\to$`let'; six tokens back, two \Delete{} actions remove `faces tried' and the model types `face trying'; four tokens back, `looks'$\to$`look'. The output is the reference sentence except for the closing `...', which the reference turns into a period. The AR rewriter writes the same sentence, so on this item the two models tie.\par
\vspace{0.3em}\textbf{Actions} on the draft.\par
{\ttfamily\footnotesize\raggedright\setlength{\parindent}{0pt}\textcolor{FigBlue}{[-{}1]} \textcolor{FigRed}{[Del]} \textcolor{FigTyrianPurple}{\textquotedbl{}~points\textquotedbl{}} \textcolor{FigBlue}{[-{}4 -{}1]} \textcolor{FigRed}{[Del]} \textcolor{FigTyrianPurple}{\textquotedbl{}~let\textquotedbl{}} \textcolor{FigBlue}{[-{}4 -{}2]} \textcolor{FigRed}{[Del]} \textcolor{FigRed}{[Del]} \textcolor{FigTyrianPurple}{\textquotedbl{}~face trying\textquotedbl{}} \textcolor{FigBlue}{[-{}4]} \textcolor{FigRed}{[Del]} \textcolor{FigTyrianPurple}{\textquotedbl{}~look\textquotedbl{}} \textcolor{FigRed}{[Stop]}\par}
\vspace{0.8em}

\section{Reproducibility Checklist}
\label[appendix]{sec:repro}

This appendix records the key artifacts and settings needed to reproduce the reported quality, trajectory, and FLOPs results.

\subsection{Artifacts and data}
\label[subappendix]{sec:repro_artifacts}

\begin{itemize}[leftmargin=*,itemsep=0.2em]
\item Code and sample outputs: in the code release for this version. It holds the training, decoding and evaluation code with the configuration file of every run; all 3,000 continuations of each model in \Cref{tab:main} (100 per model with their full action sequences) and the C4 reference continuations; every edit output of the four models in \Cref{tab:edit} on CoEdIT, IteraTeR and JFLEG; and an HTML page that replays 70 example trajectories one action at a time (also online; \Cref{sec:appendix_ranked_examples}). Its \texttt{README.md} maps each table to its scripts. The sample files reproduce the length, stop, move and delete columns of \Cref{tab:main} and the SARI and exact-match columns of \Cref{tab:edit}.
\item Repository and checkpoints: code at \url{https://github.com/Sean-Diab/Reviser} and checkpoints at \url{https://huggingface.co/sean-diab/reviser-checkpoints}. Both hold the release of the earlier version of this work; the code, sample outputs and checkpoints of this version will be added there.
\item Environment: \texttt{requirements.txt} in the code release (versions used: \Cref{sec:repro_compute}).
\item Evaluation datasets: \texttt{allenai/c4} (English validation split; 35-token prompts), DCLM (ablations), CoEdIT validation, IteraTeR human sentence test, JFLEG test; GPT-2 tokenizer.
\item Data release policy: the prepared training data and the checkpoints are not in the code release (size; the checkpoints are on Hugging Face); all scripts take local file paths.
\item Licensing and intended use: all datasets and models used are publicly available or used through the providers' standard interfaces (API or command-line tools), in accordance with their licenses, terms and intended research use. Licenses as released: C4 and FineWeb ODC-BY; DCLM-baseline and C4\_200M CC-BY-4.0; CoEdIT and IteraTeR data Apache-2.0; JFLEG CC-BY-NC-SA-4.0 (evaluation only); CoAuthor states no license (used only to fit writing-order statistics); GPT-2 MIT; SEDD MIT (code release); Pythia, Cerebras-GPT, MDLM, Dream-7B, SmolLM2-360M-Instruct, Qwen2.5-0.5B/1.5B-Instruct, Qwen2.5-32B-Instruct-AWQ and BERT-large Apache-2.0; OPT the OPT license (non-commercial); CoEdIT-large CC-BY-NC-4.0 (baseline only); Qwen2.5-3B-Instruct the Qwen Research License (RL reward only); Skywork-Critic-Llama-3.1-8B the Skywork Community License and the Llama 3.1 license (judge only); for measured FLOPs only, the LevT checkpoint MIT, the Mask-Predict checkpoint CC-BY-NC-4.0, the OpenWebText SEDD, MDLM and AR checkpoints and Qwen2.5-7B Apache-2.0, LLaDA-8B and SDLM their released licenses, and WMT14 its data terms; Codex, Claude Opus 5.5 and Claude Haiku 5.5 the providers' terms of service. Our code is released under the Apache-2.0 license. The checkpoints of this version will be released under terms that respect those of the artifacts used to train them: the +RL models were trained with a reward from Qwen2.5-3B-Instruct, and part of the Reviser-Edit training data was written by Claude Haiku. C4 and FineWeb are derived from Common Crawl web text and may contain personal or offensive content; we perform no additional data collection or de-anonymization and rely on the dataset providers' filtering, and the Ethical Considerations section notes that model outputs can reflect biases present in this data.
\end{itemize}

\subsection{End-to-end code path}
\label[subappendix]{sec:repro_codepath}

Paths are inside the code release for this version.

\begingroup\raggedright
\begin{itemize}[leftmargin=*,itemsep=0.2em]
\item Continued training: trainer \path{experiments/july_control/code/src/reviser/training/pretrain_resttraj.py}, extended by three checked patches (\path{experiments/july_control/patches/}; \path{scripts/humanorder/promptlock_patch_v1/}, the prompt lock; \path{scripts/humanorder/big608/ddp_patch.py}, multi-GPU); the released checkpoints are loaded by \path{experiments/july_control/code/july_impl/wrap_ckpt.py}. Configs: \path{configs/humanorder/big608/}, \path{configs/humanorder/big608_300m/}, and \path{configs/humanorder/overnight/} (ablations). Trajectory orders: \path{scripts/humanorder/variants.py}.
\item RL: \path{scripts/grpo/train_grpo.py}.
\item Decoding: \path{scripts/papereval/decode_batched.py} (\texttt{--lock\_prompt}); prompts: \path{scripts/papereval/build_prompts.py}.
\item Continuation metrics: \path{scripts/papereval/score_ppl.py} (GPT-2, GPT-2 Large), \path{score_dream.py}, \path{score_mauve.py}, \path{arena_skywork.py} with \path{length_control.py} (matched length), and \path{judge_cli.py} (Claude and Codex); trajectory statistics: \path{scripts/trajviz/trajstats.py}; \Delete{} probes: \path{scripts/dagger/probe_deletes.py} and \path{scripts/eval/cursor_skill.py}.
\item Editing: \path{src/reviser_v2/} (\path{train.py}, \path{evaluate_edits.py}); configs \path{configs/realedit_v3/} (seed 1) and \path{configs/realedit_v3_s2/} (seed 2); training data: \path{scripts/editdata/}; ERRANT and GLEU: \path{scripts/errant_eval.py}, \path{scripts/editextras/jfleg_gleu.py}.
\item Latency: \path{scripts/speed/bench_speed.py}, \path{scripts/speed/bench_edit_speed.py}.
\item FLOPs (\Cref{fig:efficiency}): measured, \path{scripts/benchmarks/} (\path{measure_continuation.py}, \path{fairseq_nat_flops.py}, \path{owt_family.py}, \path{large_models.py}, \path{sdlm_flops.py}; counter \path{flopcount.py}) with the records in \path{experiments/measured_flops/results.json}; analytic, \path{scripts/benchmarks/analytic_flops_released_settings.py}.
\item Commands and the table-to-script map: \path{README.md}; example viewer: \path{samples/example_viewer.html}.
\end{itemize}
\par\endgroup

\subsection{Model and training}
\label[subappendix]{sec:repro_hparams}

\begin{itemize}[leftmargin=*,itemsep=0.2em]
\item Reviser architecture: 100M uses $(L,d,h,d_{\text{ff}})=(24,512,8,2048)$; 300M uses $(26,896,14,3584)$; pre-LN, RoPE and tied embeddings. The AR baselines share these shapes.
\item Context limits: 100M $(T_{\max},L_{\max})=(255,512)$; 300M $(511,512)$.
\item Action space: moves $\Mset=\{\pm 2^k : k=0,\dots,9\}$ (offsets $\pm1$ to $\pm512$); \Delete{} (id 50259) is a backspace; \Stop{} is id 50260.
\item Released models (earlier version): AdamW ($\beta_1=0.9,\beta_2=0.95,\epsilon=10^{-8}$, wd=0.1), cosine schedule with 2000-step warmup, bf16, dropout 0.1, grad clip 1.0; learning rates 100M peak/min $10^{-4}/2\times10^{-5}$, 300M $10^{-5}/10^{-6}$; batching 100M (bs 64, seq 256), 300M (bs 16, accum 4, seq 512); obfuscation mix $p(\Delete)=0.8$, $p(\Move)=0.2$; no validity mask in training, and loss on the prompt inserts (100M: in its only stage with a prompt; 300M: for the first 5.4B of its 12.8B fine-tuning tokens).
\item Training budgets (input tokens; for Reviser these are trajectory actions, about 80\% of them inserts, so they are not directly comparable to text tokens): Reviser 100M was trained in one continuous run from random initialization for 4.83M optimizer steps (batch 64, length $\le$256) over three FineWeb restoration-trajectory stages, about 47.8B trajectory tokens in total (computed from steps and tokens per step; no counter was recorded). Reviser 300M was fine-tuned on FineWeb for 12.8B trajectory tokens from a 300M DCLM restoration model trained for about 39B tokens (its counter read 30.0B but, until step 913k, counted only one of every four micro-batches). The AR baselines were fine-tuned on FineWeb for 10.0B (100M) and 10.5B (300M) text tokens; the 300M one started from a 6.5B-token DCLM AR model, and the 100M one's starting point is not recorded. These fine-tunes were restarted many times, and each restart replayed the start of the same data order, so the Reviser 300M fine-tune saw only about 1.5B distinct trajectory tokens (over two data sets) and each AR fine-tune about 0.5B distinct text tokens.
\item Continued training (\Cref{tab:main}): from each released model, 608.7M tokens (32 sequences per step; 84,488 steps without and 79,831 with self-correction) of run-and-jump trajectories over the same 2.64M FineWeb texts, with or without self-correction (5\% near errors, immediate, masked; error seed 7); peak learning rate $2\times10^{-5}$ (100M) and $2\times10^{-6}$ (300M) with 2\% warmup and cosine decay to a tenth over a 122,070-step schedule (2B tokens at full length), which these runs stop before finishing (the rate ends about 3$\times$ above the floor); fp16 with loss scaling; data seed 123 (and torch seed 123, except the run-and-jump run without self-correction, whose dropout was unseeded); no loss on the prompt; the self-correction runs also apply the prompt-lock mask to the training loss and ran on 4$\times$RTX 4090 with DDP (2.1\,h at 100M, 4.3\,h at 300M).
\item Ablations (\Cref{tab:ablate}): same as above at 100M, stopped at 83.9M tokens, without the prompt-lock mask in the loss; two runs per RTX 3090 (\$16.32 for 34 arms).
\item RL: \Cref{sec:rl}. Reviser-Edit and the AR rewriters: \Cref{app:editdata}.
\end{itemize}

\subsection{Hardware and FLOPs accounting}
\label[subappendix]{sec:repro_compute}

\begin{itemize}[leftmargin=*,itemsep=0.2em]
\item Released models: single RTX 5090 (32GB), PyTorch 2.11 + CUDA 13. New runs: rented RTX 4090/3090 GPUs, PyTorch 2.6.0 + CUDA 12.4. Latency: one RTX 2070 SUPER (8GB), fp16 autocast, the same KV-cached decoder for Reviser and AR, 64 prompts, median of 3 passes.
\item Compute: training the released models took about 610 GPU-hours, estimated from the training logs (Reviser 100M about 123, Reviser 300M about 356 including its DCLM stage, AR 300M about 108 including its DCLM stage, and the AR 100M fine-tune about 25; the training of the AR 100M starting checkpoint was not recorded), not counting about 33 GPU-hours of restarted or rolled-back training. The new experiments of this version used about 970 rented GPU-hours, from the rental records and including setup and idle time: about 355 to write Reviser-Edit drafts with Qwen2.5, 230 for Reviser-Edit and the AR rewriters, 180 for continued training, ablations, RL and continuation evaluation, and 200 for exploratory runs not reported here.
\item FLOPs protocol for the analytic estimates: dominant-matmul proxy in \Cref{app:prior_art_efficiency_flops} with shared constants $n=128$, $|V_c|=50000$, $d=768$, $L=12$, $d_{\text{ff}}=4d$, $K_{\max}=16$ (the large models of Group 3 keep their native sizes).
\end{itemize}

\subsection{Decoding and inference configuration}
\label[subappendix]{sec:repro_decoding}

\begin{itemize}[leftmargin=*,itemsep=0.2em]
\item Continuation: legality mask $\rightarrow$ temperature 0.9 $\rightarrow$ top-$k$ 50 (ties at the cutoff kept) $\rightarrow$ multinomial; history capped at 255 positions, i.e.\ at most 220 new actions after the 35-token prompt; prompt lock on (\Cref{sec:model}). AR baselines: same sampler, 220 new tokens.
\item Prompt lock: a move is valid only if $35\le u+\Delta\le\ell$, \Delete{} only if $u>35$; the same rule is applied in decoding, in RL and in training the \Delete{} models (\Cref{sec:repro_hparams}). The decoder behind the earlier full-length results (\Cref{app:fulllength}) did not apply it.
\item Special tokens: the mask covers only moves and \Delete, so a model can also sample its start-of-input marker (or a reserved id), which the executor ignores: 316 steps in 2 of 3,000 outputs of the 100M run-and-jump + \Delete{} model (0.08\% of its actions; both repeat it to the 220-action budget), at most 0.014\% of the actions of any other model. They count toward the budget and timings, not the trajectory statistics (\Cref{tab:traj}).
\item Editing: greedy decoding (choose the action type with the largest total probability, then the arg-max within it); action caps $3|x|+64$ (edit) and $2|x|+64$ (AR); the AR rewriters emitted no \Move{} or \Delete{} action on any evaluation item.
\item Diffusion baselines: SEDD and MDLM decoded with 128 diffusion steps (stored texts of the earlier version).
\end{itemize}

\subsection{Evaluation protocol and reporting}
\label[subappendix]{sec:repro_eval}

\begin{itemize}[leftmargin=*,itemsep=0.2em]
\item Continuation prompts: 3,000 C4 validation prompts (\texttt{allenai/c4}, English; seeds 123/124/125, 1,000 each), GPT-2 tokenizer, 35-token prompt.
\item evalPPL of the continuation given the prompt: under GPT-2 base, the mean over prompts of the per-example perplexity; under Dream-7B (8 Monte-Carlo samples), the exponential of the total estimated NLL divided by the total number of continuation tokens. MAUVE: three features per text, the mean and standard deviation of its BERT-large (\texttt{bert-large-uncased}) pseudo-log-likelihoods and its token count (text including the prompt, cut to 100 GPT-2 tokens; reference: the C4 texts), as in the earlier version; mauve's default k-means seed. The standard deviation over 10 k-means seeds is 0.004--0.008; it covers the clustering only, not the sampling of texts.
\item Judges: \Cref{app:judges}. Editing metrics: SARI with one reference (mean of sentence SARI); ERRANT 3.0.0 F$_{0.5}$, single reference, edit type UNK excluded; JFLEG GLEU (our port of the official script gives the published 40.54 for the unchanged input) with the official four references, model outputs PTB-tokenized to match the tokenized sources; paired bootstrap CIs (1,000 resamples) over records.
\item Seeds: one training seed per model unless stated (RL: three at 100M; Reviser-Edit and the AR rewriters: two, the second with another data order and run seed from the same checkpoint); decoding uses fixed per-batch seeds.
\end{itemize}

\end{document}